\documentclass[pdflatex,sn-mathphys-num]{sn-jnl}

\usepackage{graphicx}%
\usepackage{multirow}%
\usepackage{amsmath,amssymb,amsfonts}%
\usepackage{amsthm}%
\usepackage{mathrsfs}%
\usepackage[title]{appendix}%
\usepackage{xcolor}%
\usepackage{textcomp}%
\usepackage{manyfoot}%
\usepackage{booktabs}%
\usepackage{algorithm}%
\usepackage{algorithmicx}%
\usepackage{algpseudocode}%
\usepackage{listings}%

\usepackage{graphicx}
\usepackage{array,multirow,multicol}
\usepackage{pgffor} 
\usepackage{pgfmath} 
\usepackage{caption}
\usepackage{subcaption}
\usepackage{adjustbox}
\usepackage{dsfont}
\usepackage{cleveref}

\theoremstyle{thmstyleone}%
\theoremstyle{thmstyletwo}%

\theoremstyle{thmstylethree}%

\begin{document}

\title[Automated Counting of Stacked Objects in Industrial Inspection]{Automated Counting of Stacked Objects in Industrial Inspection}


\author*[1]{\fnm{Corentin} \sur{Dumery}}\email{corentin.dumery@epfl.ch}

\author[1]{\fnm{Noa} \sur{Etté}}

\author[1]{\fnm{Aoxiang} \sur{Fan}}

\author[2,3]{\fnm{Ren} \sur{Li}}

\author[4]{\fnm{Jingyi} \sur{Xu}}

\author[5]{\fnm{Hieu} \sur{Le}}

\author[1]{\fnm{Pascal} \sur{Fua}}

\affil[1]{\orgname{EPFL}, \orgaddress{\city{Lausanne}, \country{Switzerland}}}

\affil[2]{\orgname{MBZUAI}, \orgaddress{\city{Abu Dhabi}, \country{United Arab Emirates}}}

\affil[3]{\orgname{Southern University of Science and Technology}, \orgaddress{\city{Shenzhen}, \country{China}}}

\affil[4]{\orgname{Stony Brook University}, \orgaddress{\city{Stony Brook}, \state{New York}, \country{USA}}}

\affil[5]{\orgname{University of North Carolina}, \orgaddress{\city{Charlotte}, \state{North Carolina}, \country{USA}}}


\newcommand{\acron}{{\it 3DC}}
\newcommand{\parag}[1]{\vspace{0.4em}\noindent\textbf{#1}}
\newcommand{\sparag}[1]{\noindent\textit{#1}}
\newcommand{\subsec}[1]{\vspace{-0.4mm}\subsection{#1}}

\abstract{
    Visual object counting is a fundamental computer vision task in industrial inspection, where accurate, high-throughput inventory tracking and quality assurance are critical. Moreover, manufactured parts are often too light to reliably deduce their count from their weight, or too heavy to move the stack on a scale safely and practically, making automated visual counting the more robust solution in many scenarios. However, existing methods struggle with stacked 3D items in containers, pallets, or bins, where most objects are heavily occluded and only a few are directly visible. To address this important yet underexplored challenge, we propose a novel 3D counting approach that decomposes the task into two complementary subproblems: estimating the 3D geometry of the stack and its occupancy ratio from multi-view images. By combining geometric reconstruction with deep learning-based depth analysis, our method can accurately count identical manufactured parts inside containers, even when they are irregularly stacked and partially hidden. We validate our 3D counting pipeline on large-scale synthetic and diverse real-world data with manually verified total counts, demonstrating robust performance under realistic inspection conditions.    
}




\maketitle

\begin{figure}[t]
    \centering
    \includegraphics[width=\textwidth]{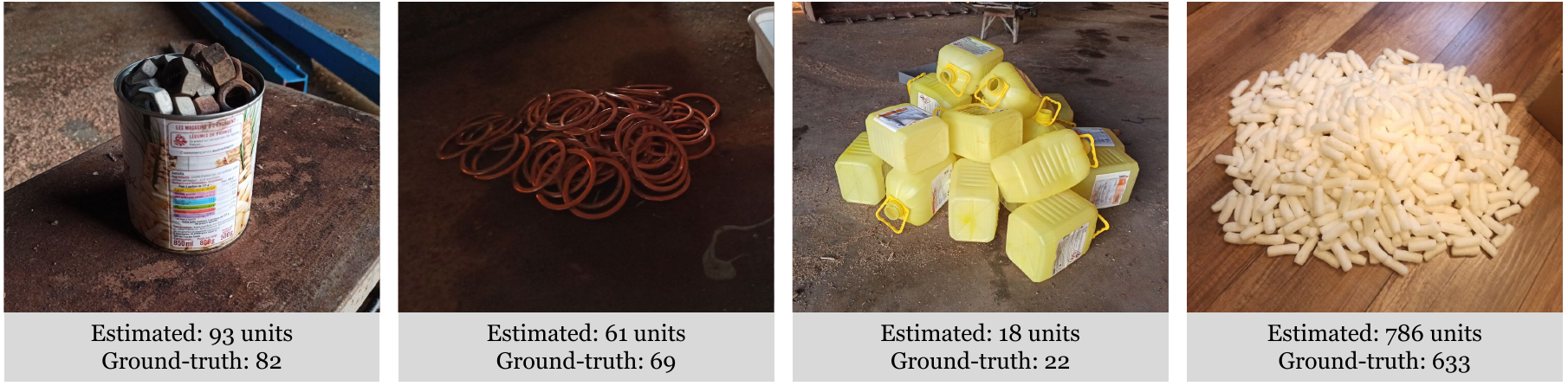}
    \caption{\textbf{3D Counting (3DC).} We estimate both the total volume occupied by the stack and the fraction of this volume taken up by the objects from multiple views of objects to be counted. Combining these estimates yields the total number of objects.}
    \label{fig:teaser}
\end{figure}

\section{Introduction}\label{sec:intro}

Object counting is a fundamental stage in numerous industrial inspection pipelines, in which manufactured items or harvested products are inspected for quality and quantity. While visual counting is a well-studied problem, most of the existing methods in the literature do not apply in this context. These methods are often specialized for specific domains such as biomedical imaging~\cite{Xie18b}, or traffic~\cite{Mandal2020} and wildlife~\cite{Arteta16}  monitoring, and general methods~\cite{Liu22a, Xu23d, Ranjan21, Shi22a,Lu18} only attempt to count objects that are directly visible in the input images, such as apples spread across a table or people in a crowd.

However, industrial inspection often involves objects that are stacked on top of each other, as in Fig.~\ref{fig:teaser}. In that case, only a subset of them is visible, making counting much more challenging for computer vision models and humans alike, as demonstrated in our experiments. However, a method to accurately and reliably count these objects from images would have significant applications in industrial and agricultural settings, where precise quantification of items would enhance operational efficiency and logistics.

In order to develop a method that can overcome the above-mentioned challenges, we will need to infer the presence of hidden objects from only limited information. As such, detecting visible object features only is not sufficient and we will also need to reason about hidden ones through contextual understanding of stacking patterns, object orientations, and irregular arrangements, making traditional counting approaches insufficient.

\begin{figure*}[htbp]
    \centering
    \includegraphics[width=\linewidth]{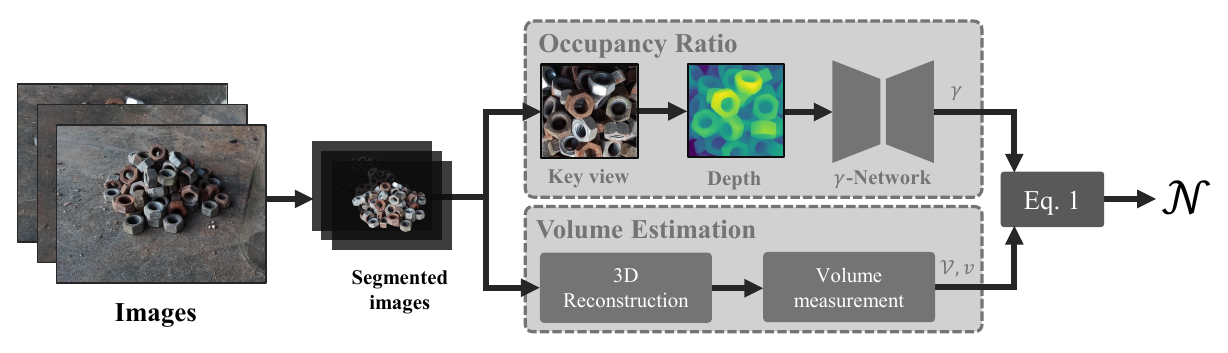}
    \caption{\textbf{3DC pipeline.} We decompose the counting task into estimating the volume of the objects to be counted and then estimating the occupancy ratio within that volume. The first is done on the basis of geometry reconstructed from segmentations in multiple images.The second uses as input a depth-map computed by a monocular depth estimator and regresses an occupancy ratio from it. }
    \label{fig:pipeline}
\end{figure*}

Our method relies on a key insight: the percentage of space occupied by objects within the container, which we will refer to as the {\it occupancy ratio}, can be accurately inferred from a depth map computed by a monocular depth estimator from a view in which objects of interest are clearly visible. This view is often one where the container is seen roughly from above, without having to be strictly vertical. To exploit this observation, we  break down the problem into two complementary tasks: estimating the 3D geometry of the object and the stack, and estimating the occupancy ratio within this volume, as depicted by~\cref{fig:pipeline}. This enables us to estimate the total count through a combination of geometric reconstruction for volume estimation and deep learning-based depth analysis for occupancy prediction. 

In our experiments, we evaluate \acron{} on a broad set of real-world and synthetic scenes. The real-world benchmark spans a variety of industrial and retail settings, with objects stacked in containers or kept in their original packaging, as illustrated in \cref{fig:teaser}. To systematically analyze the behavior and limitations of our approach, we further introduce a large-scale synthetic dataset \cite{dumery2025stackcounting} with precisely controlled configurations and ground-truth counts. Both this dataset and implementation are already released and can be found at \url{https://cvlab-epfl.github.io/projects/stacks.html}. 

This paper extends our earlier conference paper~\cite{dumery2025counting} 
to further explore the applicability of our counting method to industrial inspection, a setting in which manufactured objects are often too light or too heavy to be reliably counted using a scale.
First, we put our approach on a more solid theoretical footing by defining counts as volumetric integrals, which provides additional theoretical insights on the correctness of our approach and the assumptions we make. Second, we extend our dataset with an additional 858 images from 13 scenes captured in challenging industrial environments, increasing the diversity of our benchmark. Third, we add a more thorough evaluation of the volume estimation component of our pipeline. With this addition, we can now separately assess the effectiveness of our method at volume occupancy estimation,  volume measurement, and complete count estimation.




In short, our contributions are as follows: 
\begin{itemize}
 \item A full end-to-end framework for 3D counting of overlapping, stacked objects, addressing a previously unexplored and challenging computer vision problem.
 \item A formula to estimate the count of stacked objects, along with its formal definition and derivation.
 \item A network that predicts the occupancy ratio, a key conceptual contribution and serving as a central element of our architecture. 
 \item A large-scale 3D Counting Dataset with 400,000 images generated from 14,000 physically simulated and rendered scenes, providing accurate ground-truth object counts and programmatically computed volume occupancy.
 \item A complementary real-world benchmark comprising 3229 images from 58 scenes, acquired with reliable camera poses and manually verified total counts.
 \item A human performance study based on 1,485 annotations on real images, capturing estimates from 33 participants. 
 \item An in-depth evaluation of our predicted volume measurements, volume occupancy estimations, and total counts.  
\end{itemize}
In particular, the latter demonstrates that this task is intrinsically difficult and that even human performance is limited. This suggests that learning to predict stacked counts directly from images in a single step may be impractical and that our strategy of decomposing the problem into simpler subproblems is key to achieving strong results, as shown in~\cref{sec:experiments}.

\section{Related work}\label{sec:rw}

Previous works in visual counting seek to estimate how many instances of a particular object are present in a scene. The majority of existing methods focus on visible objects observed in a single image. In contrast, our work addresses the more demanding setting where a substantial fraction of the objects to be counted are fully or partially hidden from view. We review these relevant works below.

\parag{Single-View Counting.} Recent single-view counting techniques often train a model tailored to one target class, for example crowds~\cite{Xu19a,Shi19b,Liu19d,Sindagi19a,Cheng19a,Wan19b,Zhang19b,Zhang19c}, cars~\cite{Mandal2020}, or penguins~\cite{Arteta16}. Such approaches have been successfully applied in a wide range of domains~\cite{Chattopadhyay16}, including medical imaging for counting cells or other anatomical structures~\cite{Falk18} and remote sensing for trees or buildings~\cite{Yi23}. They tackle challenges like scale variation, perspective distortions, and partial occlusions by learning robust feature representations~\cite{Zhang16c}, predicting density maps~\cite{Sindagi17}, or exploiting multi-scale information~\cite{Ranjan18}. Beyond single-class settings, class-agnostic counting~\cite{Lu18,Nguyen22a,d2024afreeca,d2023learning} allows counting arbitrary categories at test time from a few support examples~\cite{Yang21f}, a handful of bounding boxes~\cite{Ranjan21}, or even just a class name~\cite{Xu23d,amini2023open,amini2024countgd,amini2025open}. Similar to these class-agnostic approaches, our method is not tied to particular object categories and can generalize across types. However, almost all of these methods assume that the instances are visible. The closest work to ours is~\cite{Jenkins23}, which leverages LiDAR to infer counts of occluded objects on retail shelves for a fixed set of beverage categories with known volumes, such as ``\textit{Coca-Cola 20oz bottle}'' or ``\textit{milk carton}''. In contrast, our approach does not rely on LiDAR and is applicable to a broader variety of scenes and object types. 

\parag{Multi-View Counting.} Methods that operate on multiple views typically aim to improve counting accuracy by aggregating information across camera viewpoints. A common strategy is to project feature maps onto a shared ground plane to obtain accurate density maps for crowd counting~\cite{Zhang20g,Zhang20b} or segmentation maps for fruit counting~\cite{Nellithimaru19}. Despite using multiple views, these approaches still presuppose that each object is visible in at least one camera and usually target a specific object class, which limits their applicability in more complex environments such as industrial inspection. Our work instead focuses on the largely unexplored challenge of estimating counts for heavily occluded objects, while placing no constraints on their category or geometry.

\section{Method}\label{sec:method}

Estimating the total number $\mathcal{N}$ of objects in a container from 2D images alone represents a fundamental challenge in industrial inspection, one that remains difficult even for human operators and has not been addressed by prior computer vision methods. The core principle underlying our method is that, despite our inability to precisely reconstruct the 3D configuration of all occluded objects at the bottom of the container, the occupancy ratio can still be reliably inferred from a single image when a sufficient number of objects are visible. This observation is particularly relevant in industrial settings where containers are typically viewed from above, providing satisfying visibility of the top layers. 

This section begins by formalizing the stack counting problem in~\cref{sec:statement}, then we mathematically derive the stack counting formula in~\cref{sec:formula}. We describe our occupancy ratio estimation network in~\cref{sec:occupancy}, and finally present our volume estimation algorithm in~\cref{sec:volume_estimation}.

\subsection{Problem Statement}
\label{sec:statement}

\parag{Task setting.}
Specifically, the task of multi-view stack counting takes as input a set of images $\{I_i\}_{i=1}^{N_I}$ and outputs a single scalar $\mathcal{N}$ representing the total count. We first localize the region of interest by segmenting the objects $S_{o}$ and their container $S_{c}$ in an initial frame, and leverage SAM2~\cite{Ravi24a} to propagate these segmentations as $S_{o,f}$ and $S_{c,f}$ across all subsequent frames $f$, enabling automatic identification of the frame with the best visibility of the objects. Finally, in non-industrial applications where the volume of a single object $V_o$ may not be known, it is estimated with the method of Sec.~\ref{sec:volume_estimation} from a separate set of images. 

\parag{Assumptions.}
We make two key assumptions: objects are stacked uniformly in bulk and are approximately identical in size and shape. Additionally, we require that a sufficient number of objects remain partially visible, enabling reliable estimation of the occupancy ratio $\gamma$ from images.

\parag{Applicability.}
These assumptions are satisfied in the real-world industrial scenes shown in ~\cref{fig:teaser} and are sufficiently general to apply across diverse industrial inspection contexts. 
In retail and logistics, our approach enables automated inventory auditing by accurately counting stacked items without manual intervention, streamlining restocking operations and reducing manual labor. 
For manufacturing quality assurance, our method ensures that production batches contain the correct number of components before shipping, preventing costly discrepancies and improving supply chain reliability. 
Beyond counting, the 3D scene understanding provided by our method supports autonomous robotic systems performing pick-and-place operations in industrial environments containing such large quantities of objects.

\subsection{Counting as Integration}
\label{sec:formula}

\begin{figure}[t]
    \centering
    \setlength{\tabcolsep}{4pt}
    \renewcommand{\arraystretch}{1.0}
    \begin{tabular}{ccccc}
        Object & Stack geometry & Sliced view & Occupancy & Image \\
        \includegraphics[width=0.17\textwidth]{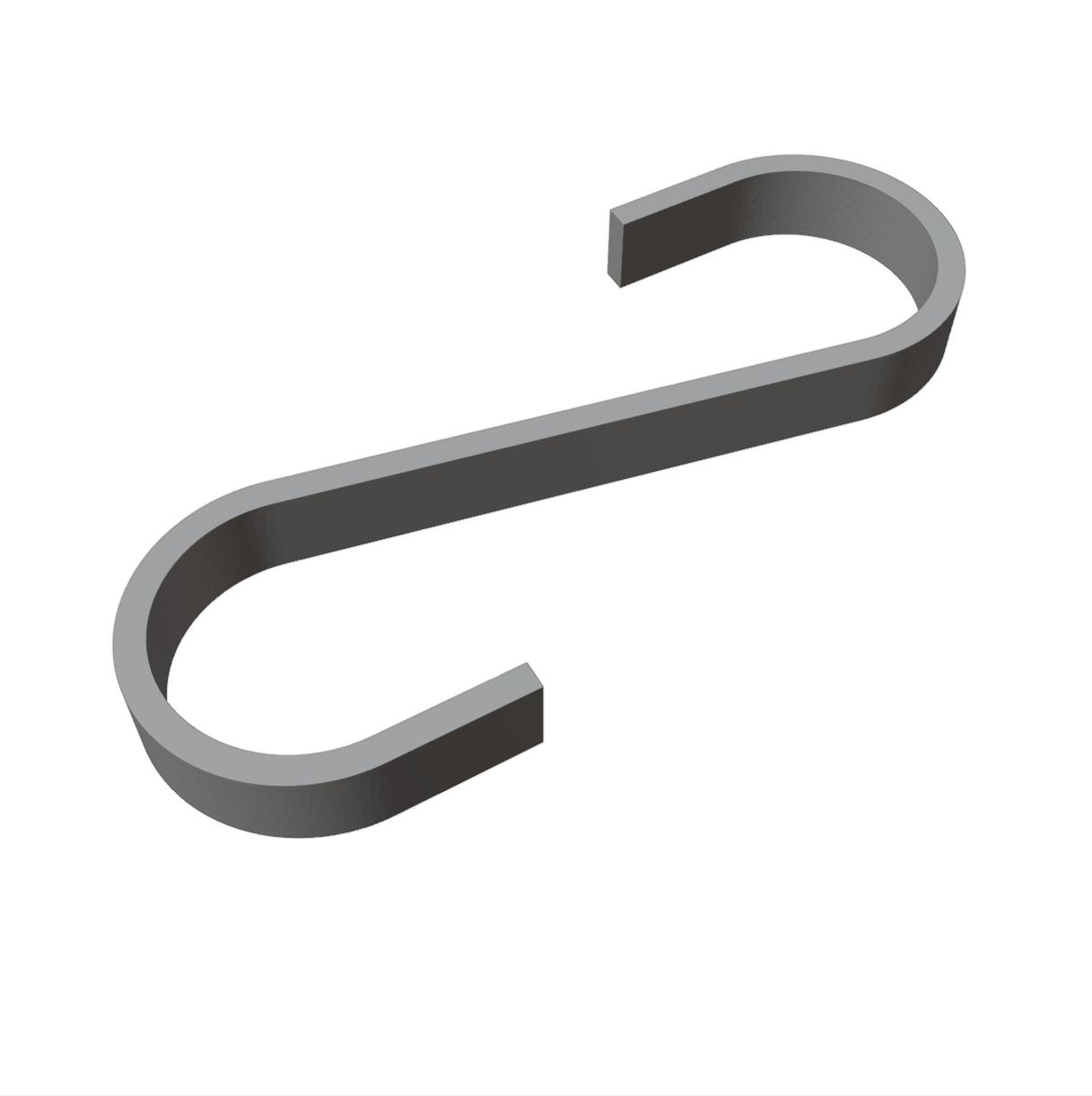} &
        \includegraphics[width=0.17\textwidth]{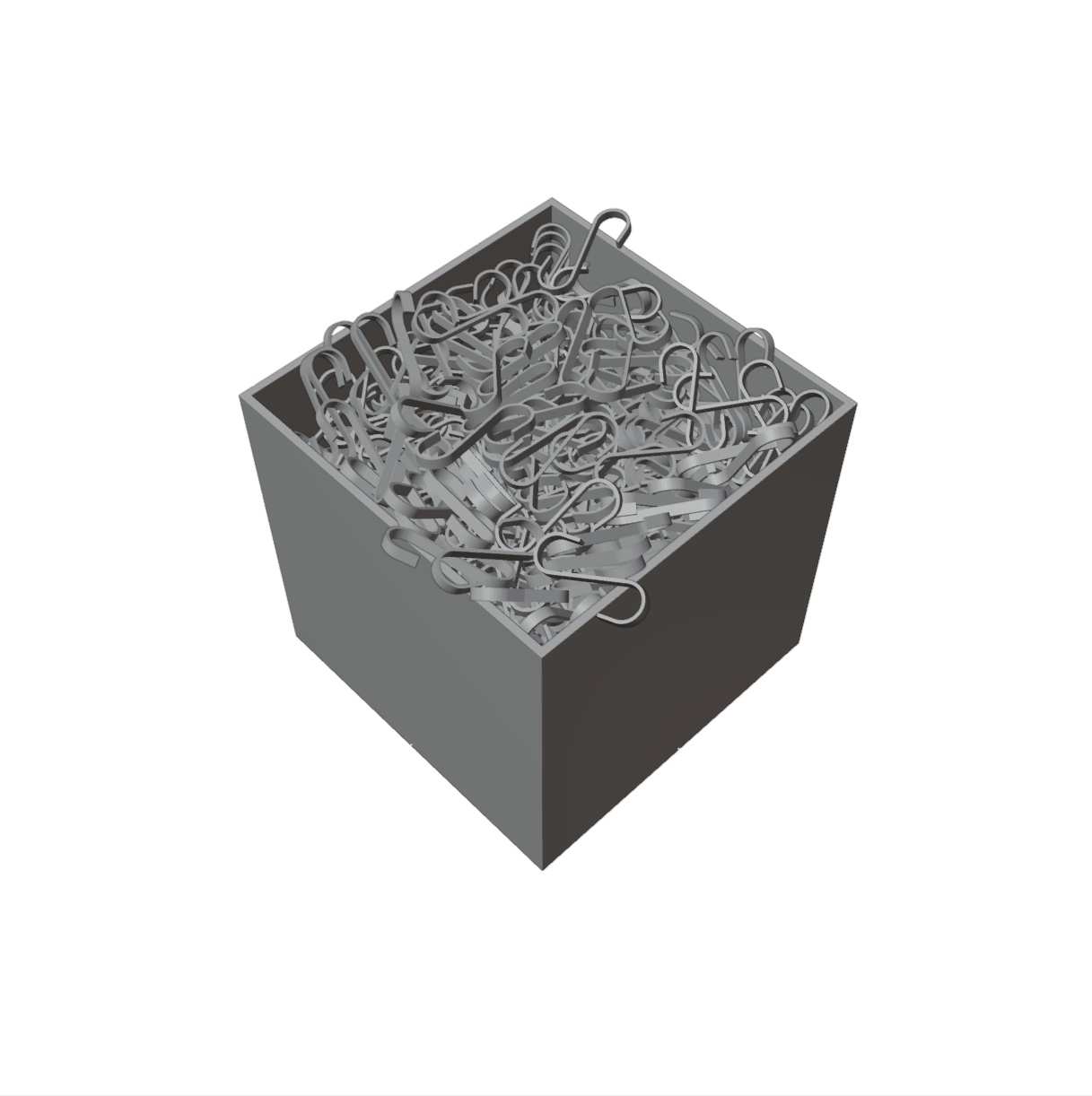} &
        \includegraphics[width=0.17\textwidth]{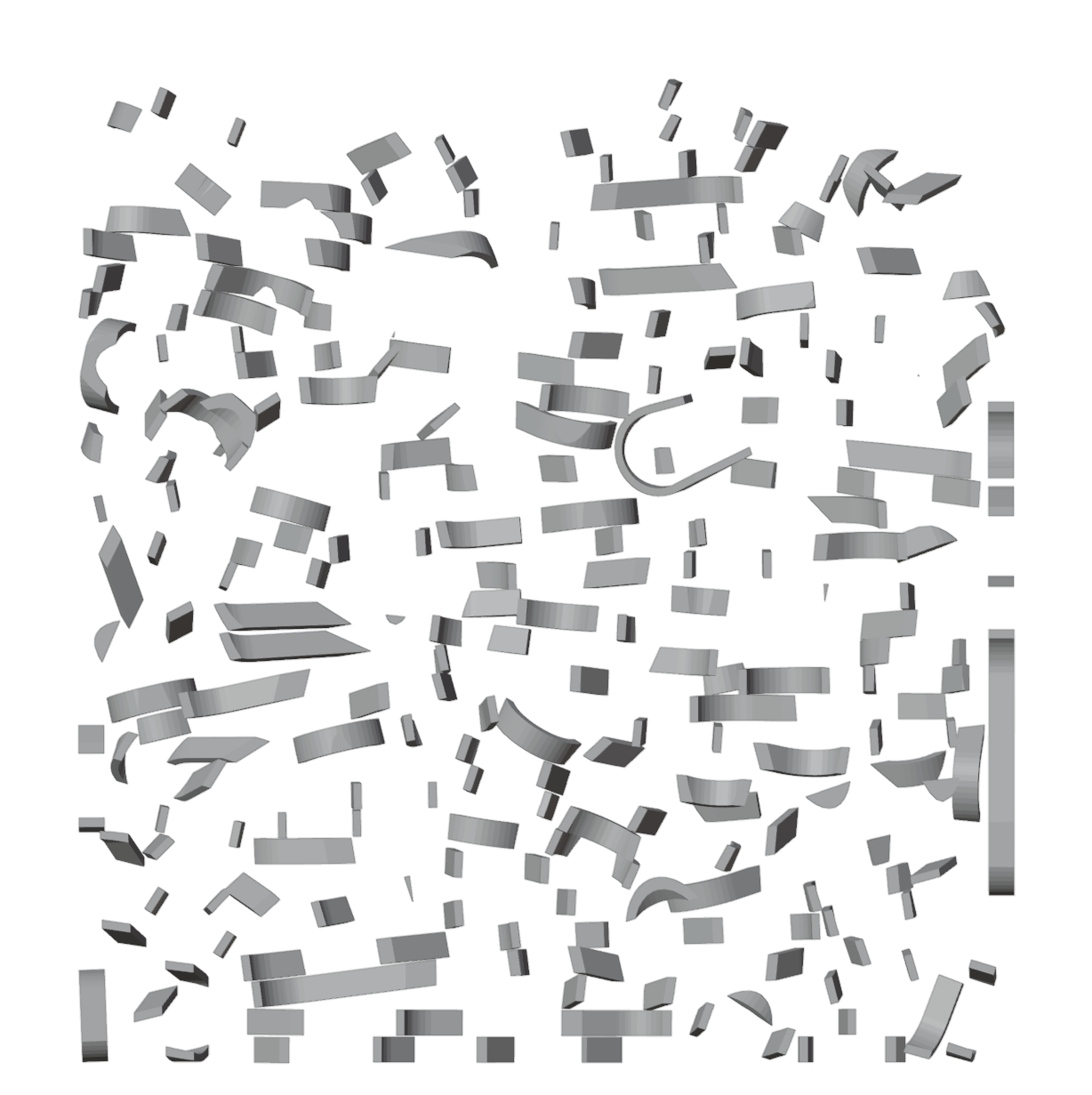} &
        \includegraphics[width=0.17\textwidth]{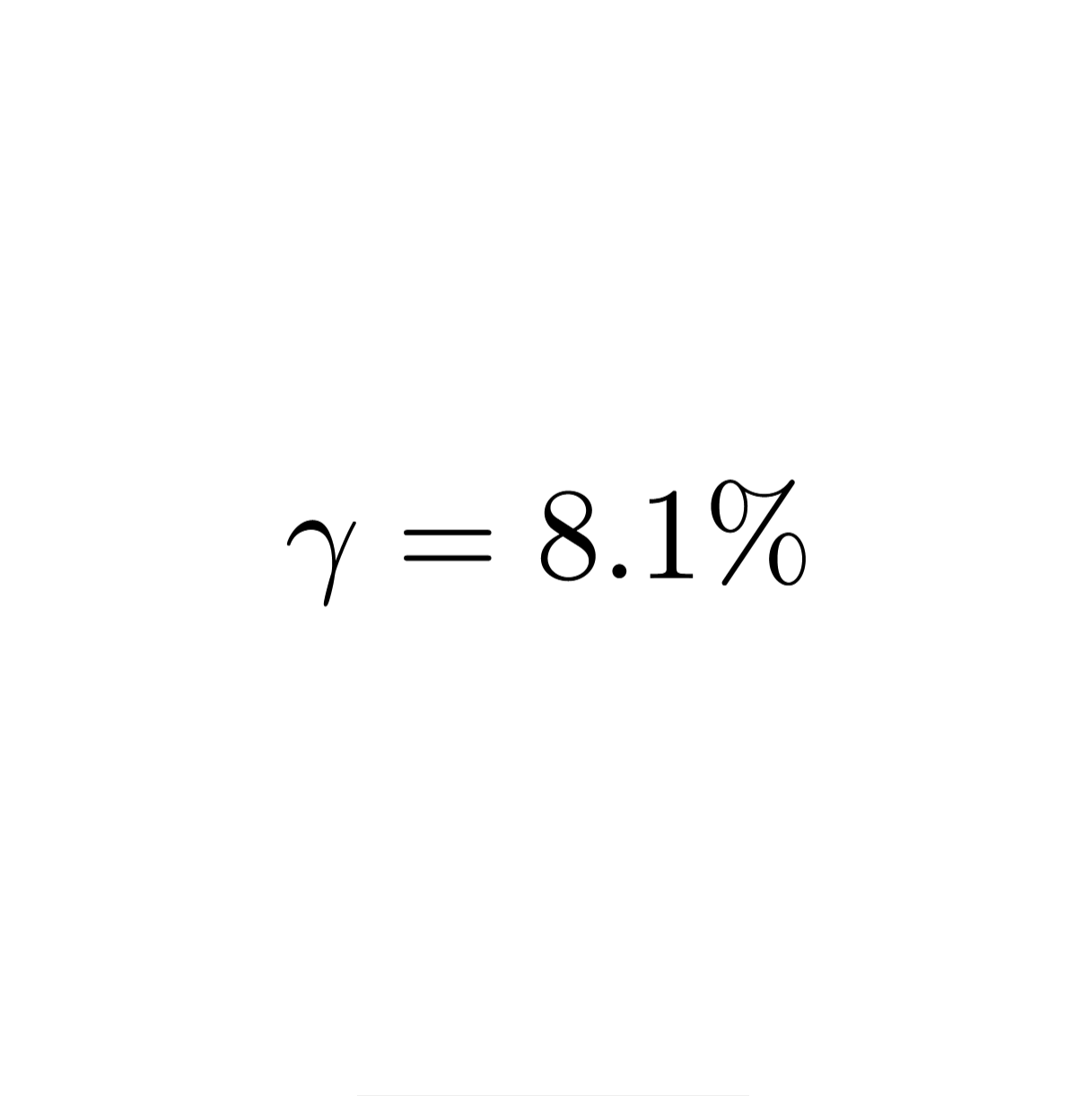} &
        \includegraphics[width=0.17\textwidth]{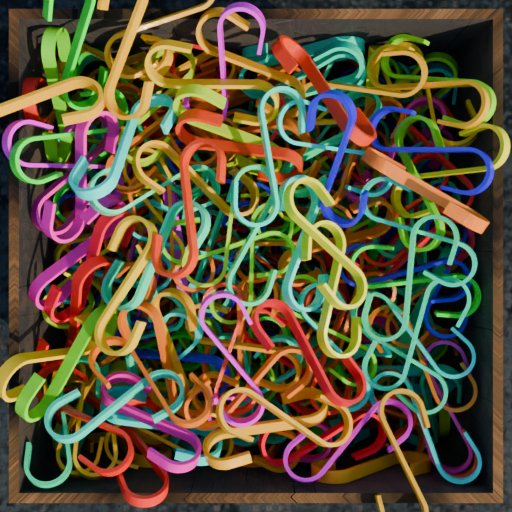} \\
        \includegraphics[width=0.17\textwidth]{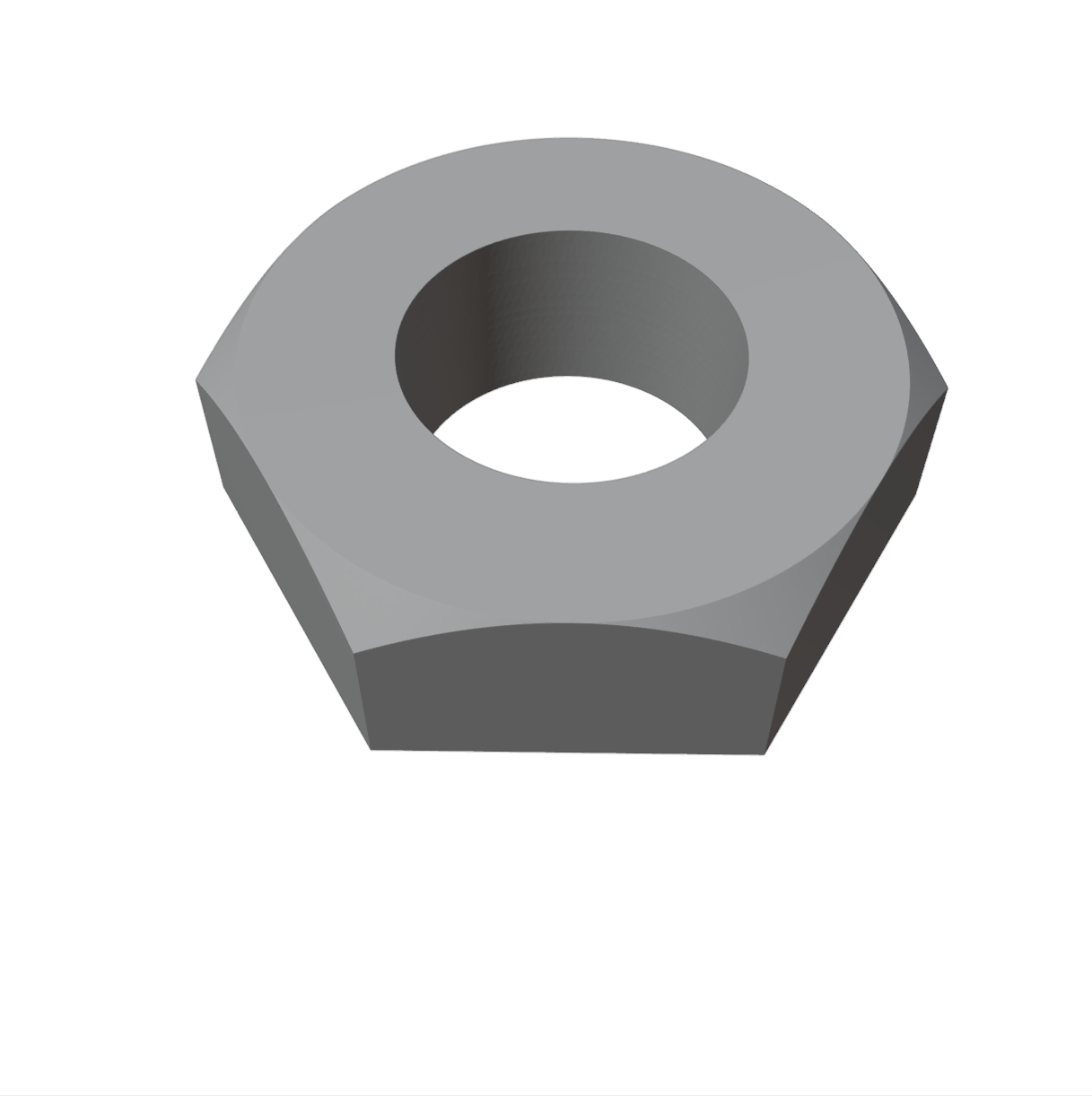} &
        \includegraphics[width=0.17\textwidth]{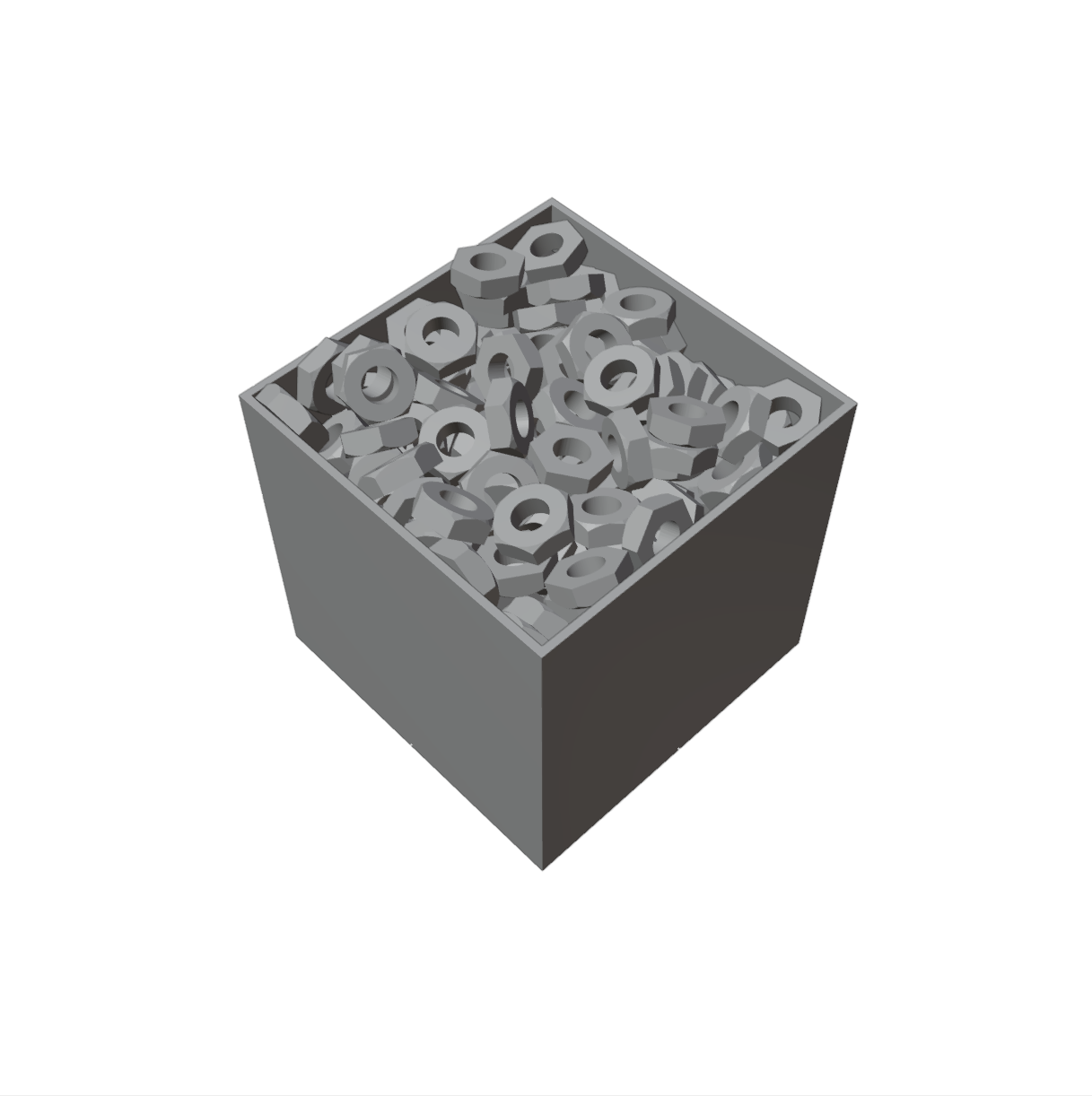} &
        \includegraphics[width=0.17\textwidth]{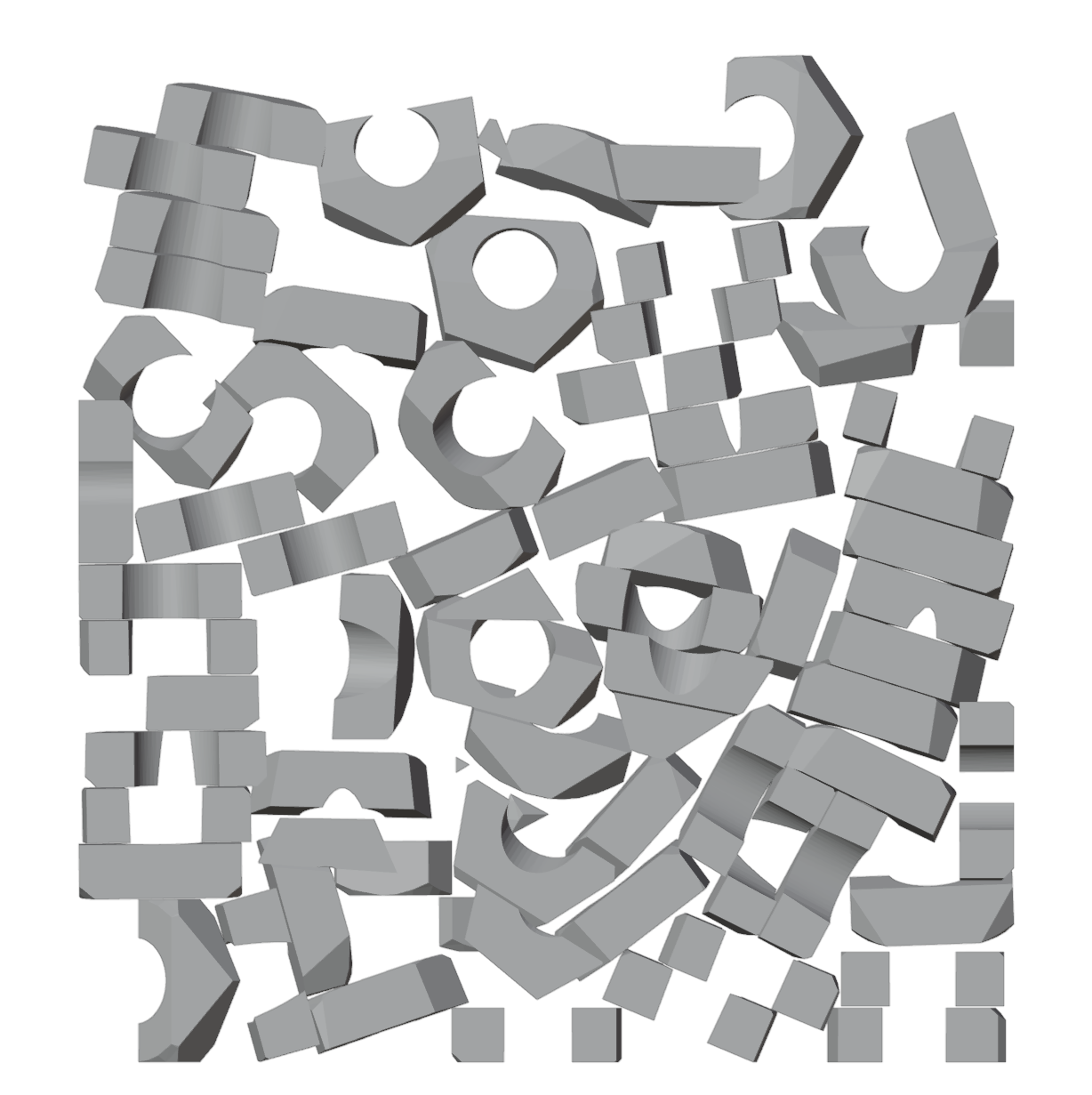} &
        \includegraphics[width=0.17\textwidth]{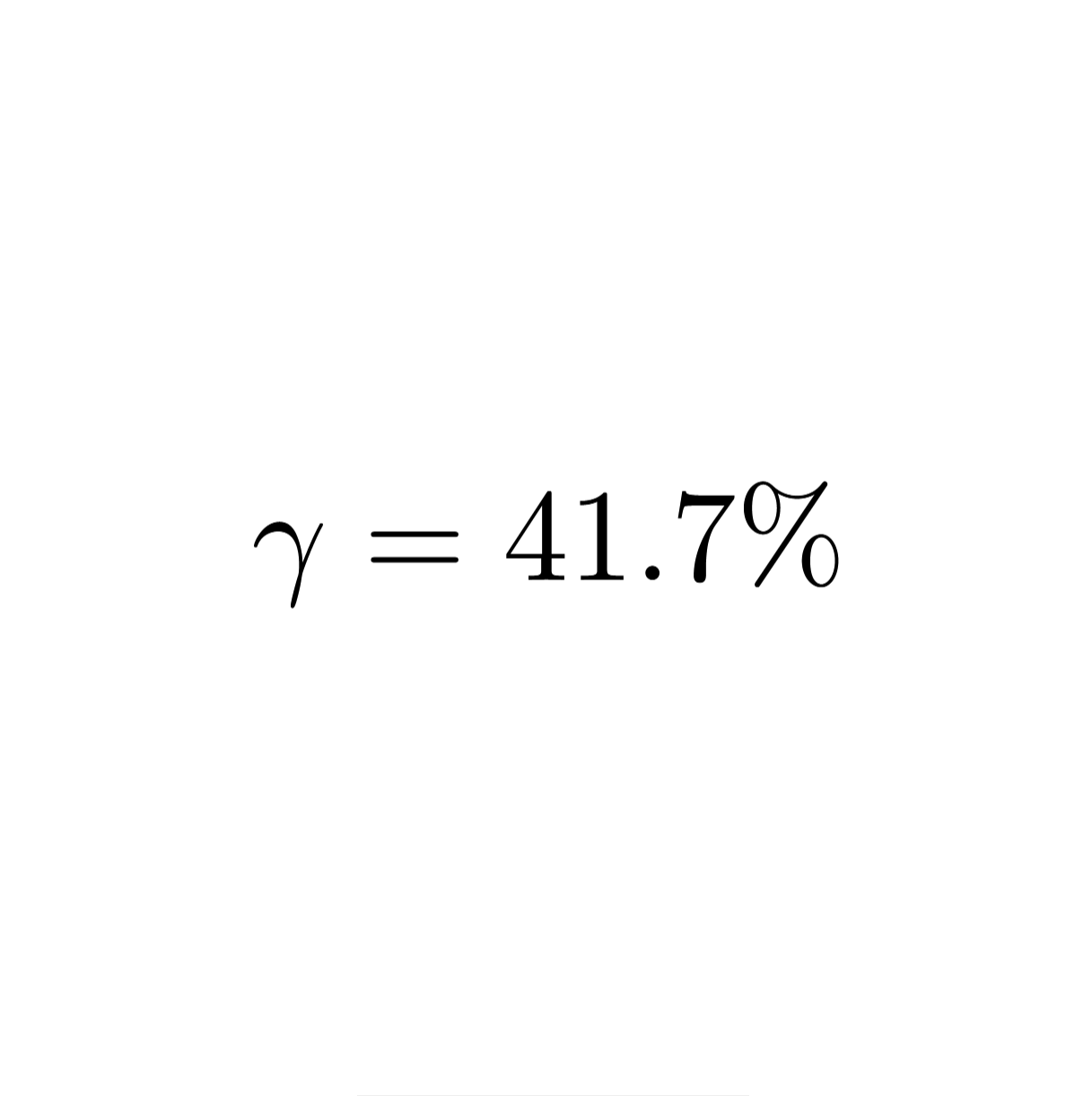} &
        \includegraphics[width=0.17\textwidth]{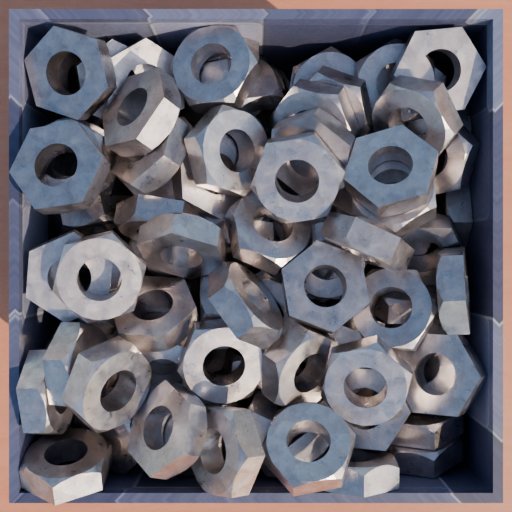} \\
        \includegraphics[width=0.17\textwidth]{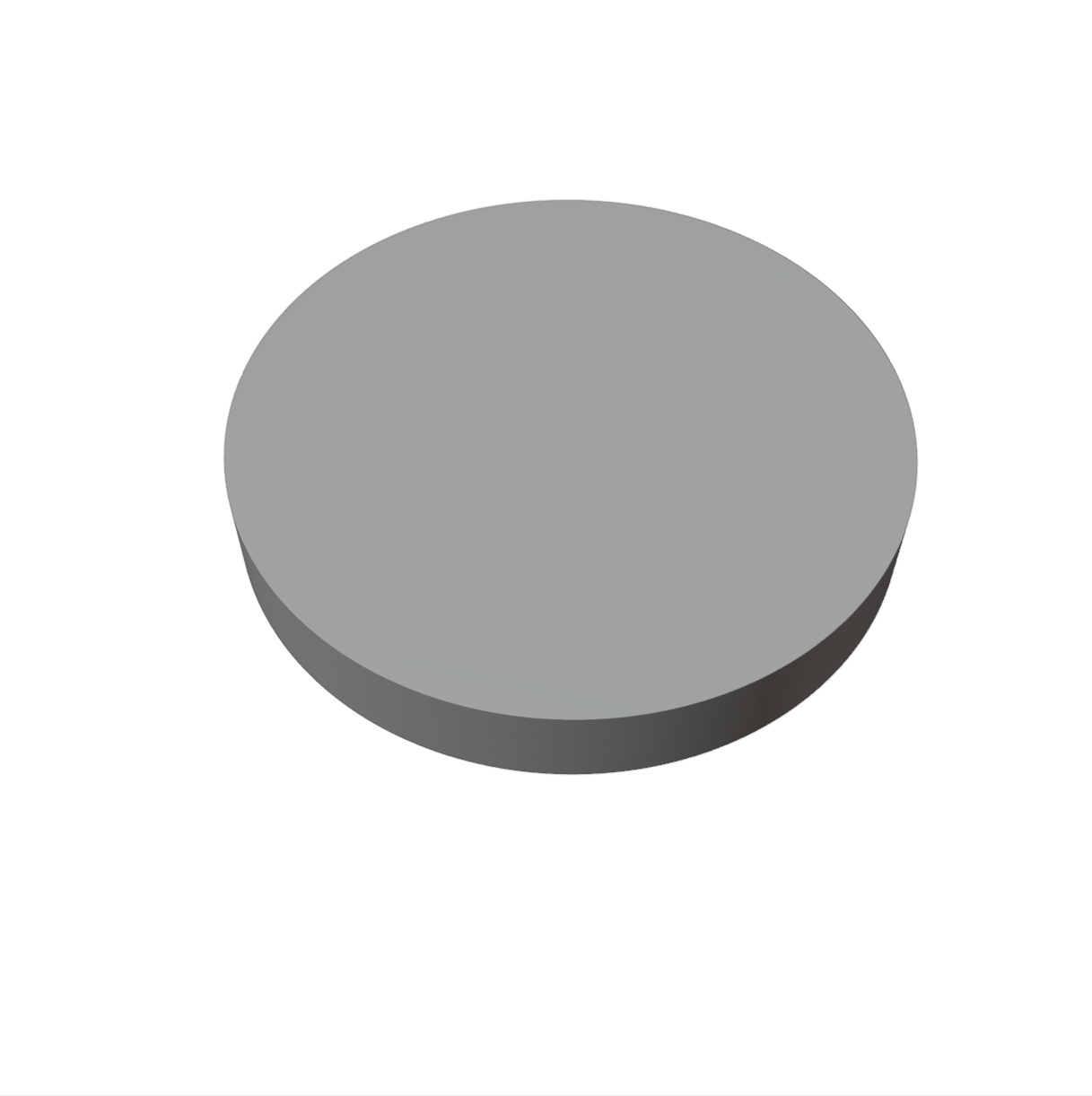} &
        \includegraphics[width=0.17\textwidth]{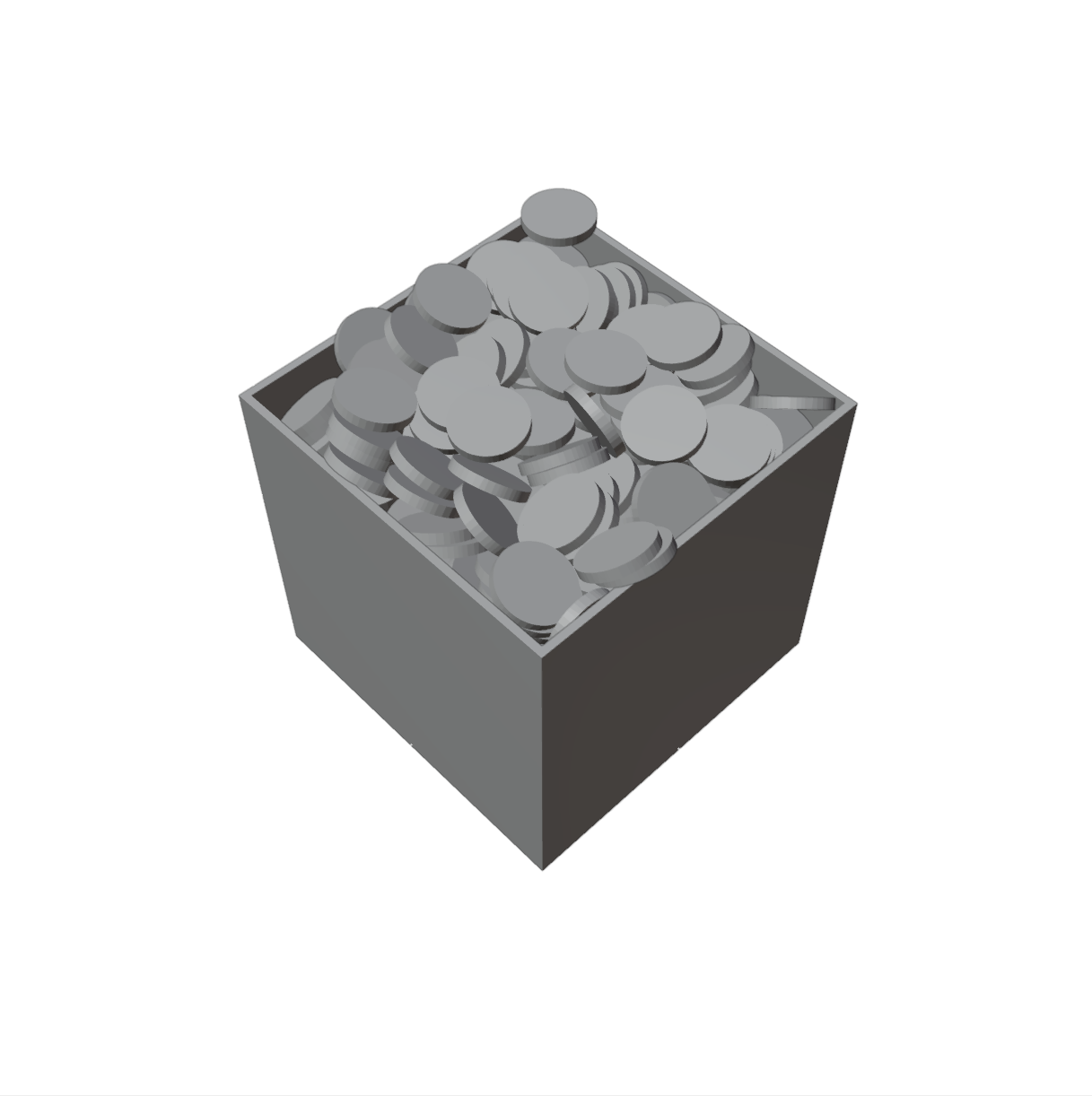} &
        \includegraphics[width=0.17\textwidth]{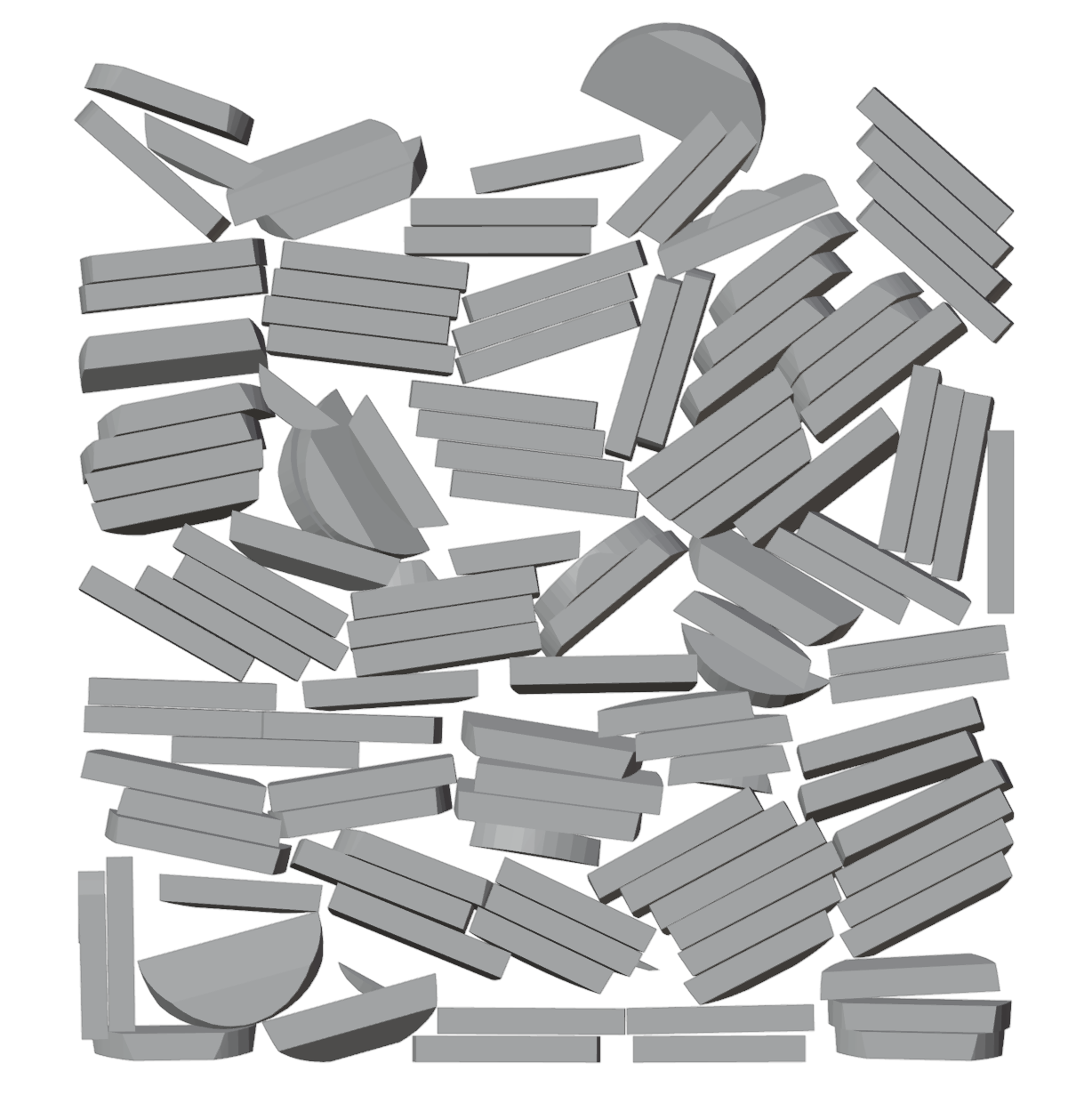} &
        \includegraphics[width=0.17\textwidth]{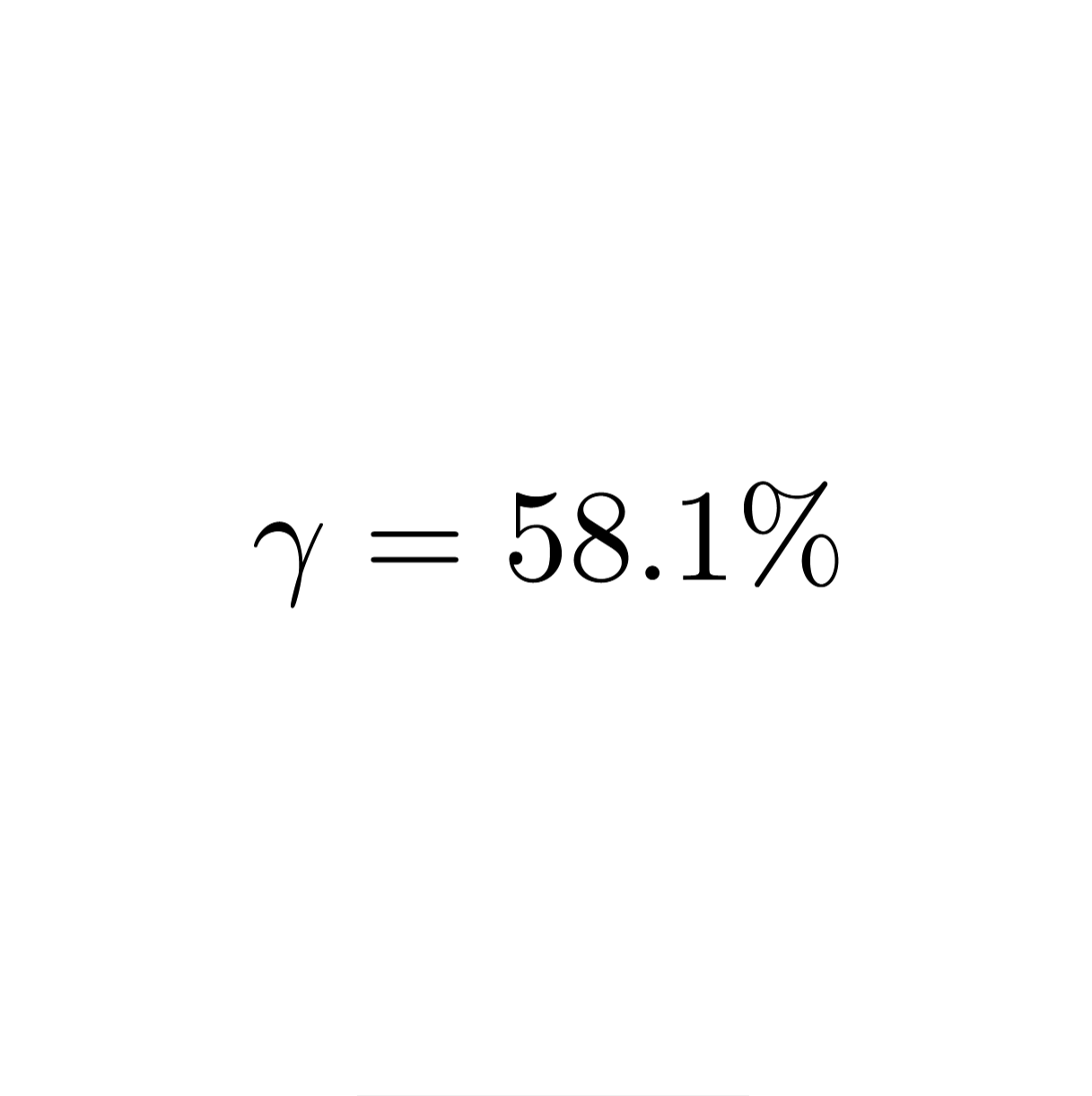} &
        \includegraphics[width=0.17\textwidth]{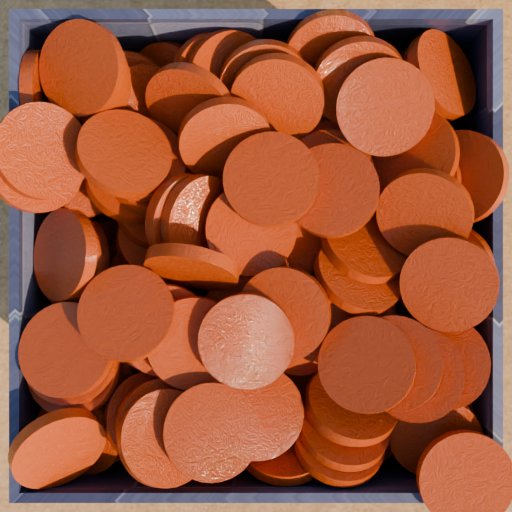}
    \end{tabular}
    \caption{\textbf{Shape-dependent volume occupancy.} We show a slice of the stacked objects for varying shapes. Different shapes yield different configurations and the fraction of space occupied by the gaps between objects varies. This is reflected in the images, where deeper layers remain visible for low $\gamma$ values.}
    \label{fig:slices}
\end{figure}

Given a stack of approximately identical objects $o$, let the volume of a single one be $V_o$.  Let $\mathcal{C}$ be a container holding the total volume $\mathcal{V}$  spanned by all the objects, which we take to be the convex hull formed by the set of all stacked objects.  
One could try estimating the number of objects as  $\mathcal{N} = \mathcal{V} / V_o$, the ratio of the total volume to the volume of a single object. However, this would be too large in general because the container is rarely completely filled. In most cases, there are substantial gaps between the stacked objects, especially in industrial bulk handling scenarios where items fall from a conveyor belt into a shipping box or truck. Thus,
$\mathcal{V}$ comprises the volumes of all objects plus the gaps between them, and we need a more sophisticated analysis.

Let $\mathds{1}(x, y, z)$ be the indicator function that denotes the presence or absence of an object at a 3D point $(x, y, z)$. The volume truly occupied by the union of all objects can be written as
\begin{align}\label{eq:integral}
\iiint_{\mathcal{C}} \mathds{1}(x,y,z) \, dx\,dy\,dz \leq \mathcal{V} \; .
\end{align}
It is naturally smaller than $ \mathcal{V}$ due to the gaps between objects. The total number of objects $\mathcal{N}$ can then be obtained by dividing it by $V_o$, yielding
\begin{align}\label{eq:exact_n}
\mathcal{N} = \frac{1}{V_o}\iiint_{\mathcal{C}} \mathds{1}(x,y,z) \, dx\,dy\,dz \; .
\end{align}
Unfortunately, there is no easy way to evaluate the volume integral, especially when the objects are stacked and those at the bottom cannot be seen. Instead, we introduce a quantity that {\it can} be evaluated from images, the volume occupancy ratio
\begin{align}\label{eq:gamma}
\gamma = \frac{1}{\mathcal{V}}\iiint_{\mathcal{C}} \mathds{1}(x,y,z) \, dx\,dy\,dz \; ,
\end{align}
which is the ratio of the volume taken up by the objects to the total volume. In this work, we assume $\gamma$ to be roughly constant across $\mathcal{V}$. While this is not strictly true, local  fluctuations tend to average in sufficiently large containers and our experiments confirm this to be the case. In the remainder of the paper, we will show that under this assumption $\gamma$, unlike the volume of Eq.~\ref{eq:exact_n}, can be estimated from the appearance of the top of the stack as seen in the images. 

Finally, combining~\cref{eq:exact_n} and~\cref{eq:gamma} yields
\begin{align}\label{eq:counting_eq}
\mathcal{N} = \frac{\gamma  \mathcal{V}  }{V_o} \; , 
\end{align}
which means that $\gamma$ and $\mathcal{V}$ can be estimated independently, as discussed in the following sections.

%
%
%
%
%

\subsection{Estimating the Volume Occupancy Ratio $\gamma$}
\label{sec:occupancy}

Central to our approach is estimating the occupancy volume ratio $\gamma$ of Eq.~\ref{eq:counting_eq} from a single image where enough target objects are visible. This image is typically taken to be the top-down view, though this is not strictly required. 
 
Concretely, we seek to learn a function $\Phi: \mathcal{D} \rightarrow \gamma$ that takes as input a depth map and outputs an occupancy ratio $\gamma \in \ensuremath{[0,1]}$, i.e., the percentage of occupied volume. $\Phi$ captures how depth maps correlate with the occupancy ratio: when deeper layers of the stack are still visible, larger gaps exist and the occupied volume fraction is lower. This dependency is agnostic to the precise object shape and should generalize to unseen geometries, as demonstrated in our experiments.

\parag{Network Architecture.} We implement $\Phi$ with an encoder architecture that first extracts rich image features and then maps them to our target $\gamma$. For the encoder pretraining, we rely on DinoV2~\cite{Oquab23}, a foundation model trained on diverse real-world imagery to improve generalization and speed convergence. 
To collapse feature maps into a single scalar for the whole image, we stack convolutional layers with ReLU activations to progressively reduce resolution, shrinking the encoded feature map to one pixel with 64 channels, and use a final linear layer to predict the scalar output. Further architectural details are provided in the supplementary material.

\newcommand{\inputnames}{00006132,00009652,00003960,00008958,00008316,00009039}
\newcommand{\labels}{{2.9, 18.9, 29.4, 38.8, 49.3, 63.7}}

\begin{figure*}[ht]
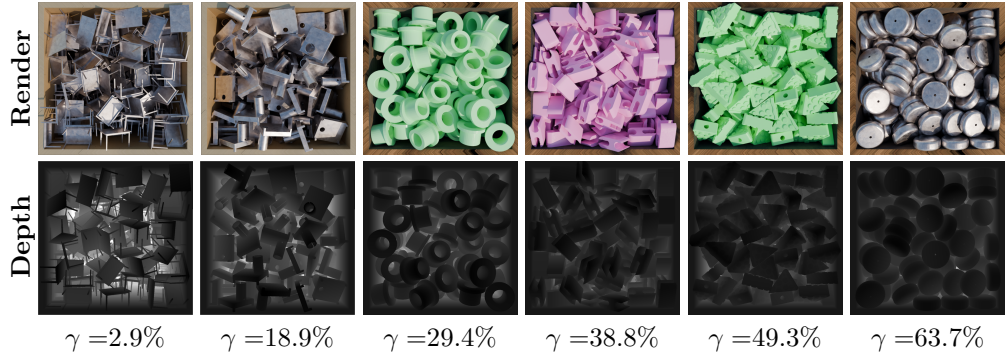

    \centering
    \begin{minipage}[c]{0.01\textwidth}
        \centering
        \rotatebox{90}{\textbf{Render}}
    \end{minipage} \hspace{-0.27em}
    \foreach \input [count=\i from 0] in \inputnames {
        \begin{minipage}[c]{0.155\textwidth} 
            \centering
            \includegraphics[width=\textwidth]{images/samples/\input_RGB.png}
        \end{minipage}%
    }

    \vspace{0.15em}
    \begin{minipage}[c]{0.01\textwidth}
        \centering
        \rotatebox{90}{\hspace{1.5em}\textbf{Depth}}
    \end{minipage}
    \foreach \input [count=\i from 0] in \inputnames {
        \begin{minipage}[c]{0.155\textwidth} 
            \centering
            \includegraphics[width=\textwidth]{images/samples/\input_DepthGT.png} \\
            $\gamma = $\pgfmathparse{\labels[\i]}\pgfmathresult \% 
        \end{minipage}%
    }
    
    \caption{\textbf{Dataset samples.} We visualize generated scenes in ascending order of occupancy ratio, with ground-truth depth maps.}
    \label{fig:samples}
\end{figure*}

\parag{Training Data.} For training $\Phi$, we minimize the squared error between predicted and ground-truth occupancy ratios. Because no suitable dataset exists, we synthesize one comprising 400,000 images across 14,000 scenes with varied objects and containers. Fig.~\ref{fig:samples} depicts some examples from our released dataset. 

To generate it, we employed the ABC dataset~\cite{Koch19a} that features a large variety of \textit{computer-assisted design} (CAD) models. We discard meshes that are not watertight or have several connected components, and rescale remaining objects to fit into a cube of side $0.05$.
Then, we initialized a virtual 3D scene with a container, and used a physics-based simulator to drop an initial batch of 100 identical objects in that box. This step was repeated a random number of times or until the box was full and the union of objects reached the space above the box after the physical simulation has converged. In each scene, the container is given a random shape and scale. Some scenes were also generated without any container where objects are directly stacked on the floor, and some scenes were generated with boxes that are partially full, as is often the case in industrial inspection. 

\begin{figure}[t]
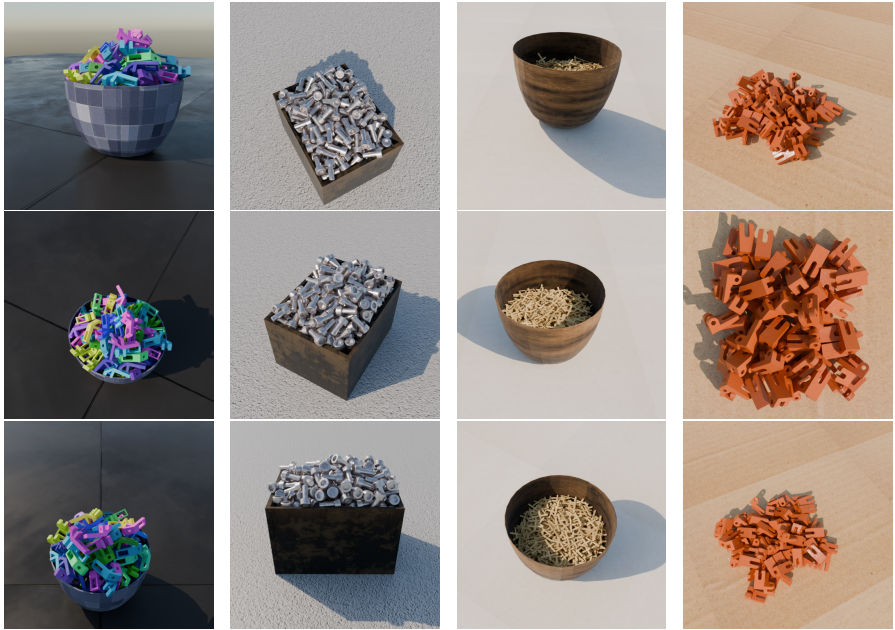

    \centering

    \foreach \img in {1_0, 3_0, 2_0, 4_0} {
        \begin{subfigure}{0.21\columnwidth}
            \centering
            \includegraphics[width=\linewidth]{images/multiview/\img.png}
        \end{subfigure}
    }    

    \foreach \img in {1_1, 3_1, 2_1, 4_1} {
        \begin{subfigure}{0.21\columnwidth}
            \centering
            \includegraphics[width=\linewidth]{images/multiview/\img.png}
        \end{subfigure}
    }

    \foreach \img in {1_2, 3_2, 2_2, 4_2} {
        \begin{subfigure}{0.21\columnwidth}
            \centering
            \includegraphics[width=\linewidth]{images/multiview/\img.png}
        \end{subfigure}
    }

    \caption{\textbf{Multi-view images.} We generate 30 views from arbitrary angles for each of the scenes in our large-scale synthetic dataset.}
    \label{fig:multiview}
\end{figure}

After the physical simulation completed, we analytically computed the occupancy ratio and the total number of objects in the container. For each object and container, we randomly assigned a realistic material that could be metallic, wood, or plastic. Some scenes also assigned a different material for each individual object. In the last stage, we used a ray-tracing engine to render multiple realistic images of the container and objects seen from several different angles to allow 3D reconstruction, as shown  in \cref{fig:multiview}. Importantly, we use ray tracing such that the bottom of the boxes appears darker due to ambient occlusion. 

This is repeated for over 14,000 scenes, and we set aside a subset of 100 scenes to use as a test set of shapes unseen during training. We generate a scene with a unit-cube container to reliably measure the ground truth $\gamma$ for each one of the 4800 shapes used, as well as two additional scenes with random containers.  With this dataset, we are able to not only to train $\Phi$ on the top view, but also to run our complete pipeline on the multi-view images in order to measure the accuracy of our count estimate $\mathcal{N}_{est}$, as performed in \cref{sec:experiments}. While our model is trained on depth maps produced by the depth estimator, our dataset also includes ground-truth depth maps, which we employ in an ablation study in our experiments to assess the requirement of accurate depth maps of our method. Additional statistics that highlight the diversity of the proposed dataset are reported in \cref{fig:dataset_stats} of our supplementary.

\begin{figure}[t]
    \centering

    \hfill
    \begin{subfigure}{0.24\columnwidth}
        \centering
        \includegraphics[width=\textwidth]{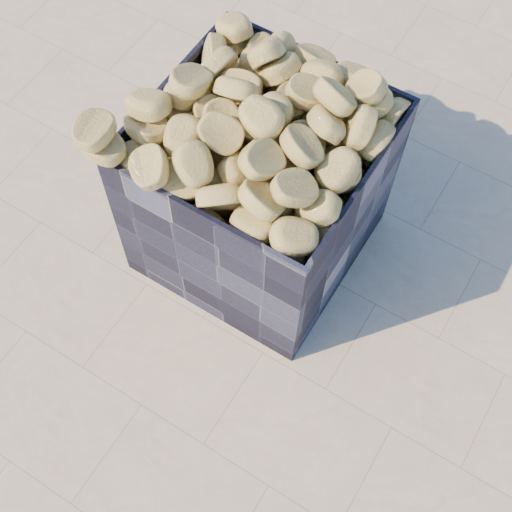}
    \end{subfigure}
    \begin{subfigure}{0.24\columnwidth}
        \centering
        \includegraphics[width=\textwidth]{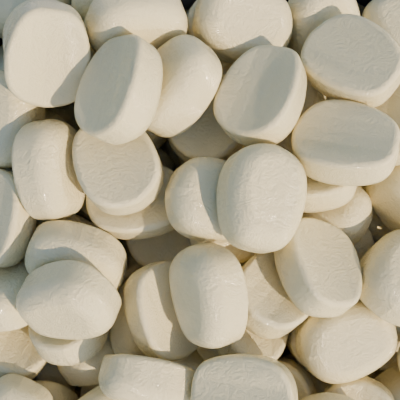}
    \end{subfigure}
    \begin{subfigure}{0.24\columnwidth}
        \centering
        \includegraphics[width=\textwidth]{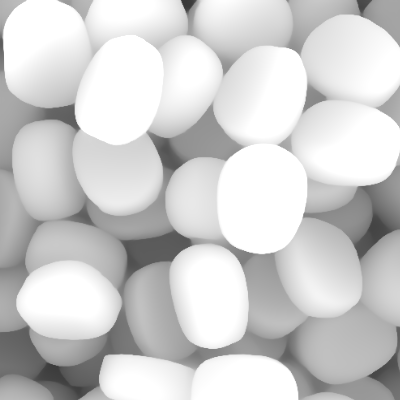}
    \end{subfigure}
    \hfill
    \vspace{0.3em}

    \hfill
    \begin{subfigure}{0.24\columnwidth}
        \centering
        \includegraphics[width=\textwidth]{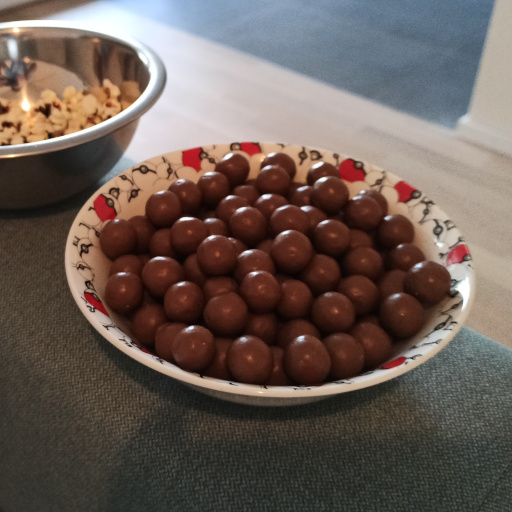}
        \subcaption*{(a)}
    \end{subfigure}
    \begin{subfigure}{0.24\columnwidth}
        \centering
        \includegraphics[width=\textwidth]{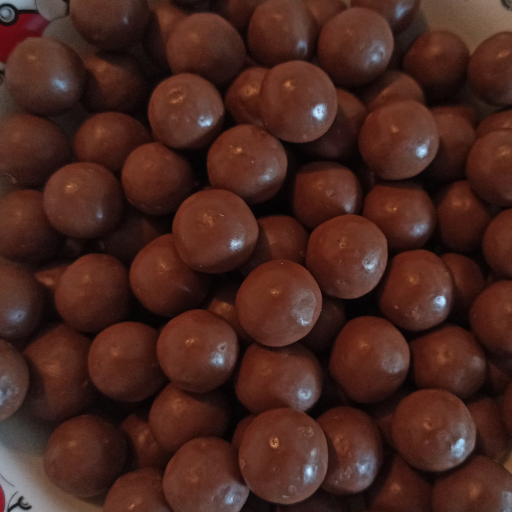}
        \subcaption*{(b)}
    \end{subfigure}
    \begin{subfigure}{0.24\columnwidth}
        \centering
        \includegraphics[width=\textwidth]{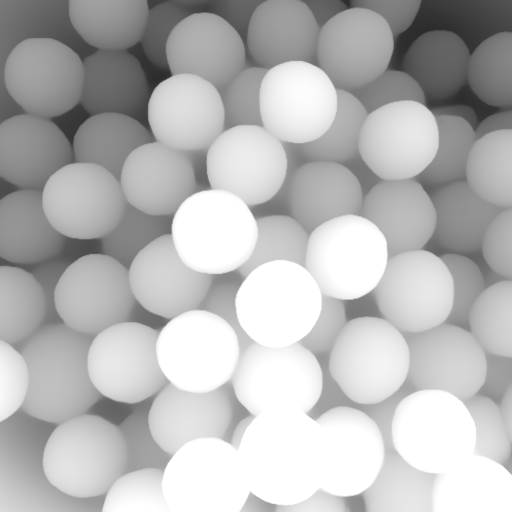}
        \subcaption*{(c)}
    \end{subfigure}
    \hfill
    \vspace{-0.3em}
    \caption{\textbf{Reducing the domain gap.} Instead of estimating the occupancy ratio $\gamma$ from synthetic (top) and real images (bottom) (a), we identify a \textit{key view} (b) and train a network to predict $\gamma$ from their depth maps (c), which are indistinguishable. 
    Top row: synthetic, $\gamma_{gt} = 62.4\%$. Bottom row: \textit{chocolates}, $\gamma_{est} = 53.5\%$, $\mathcal{N}_{est} = 119$, $\mathcal{N}_{gt} = 131$.}
    \label{fig:similarity}
\end{figure}

\parag{Inference.} When applying our method on real data, we must pick which image to feed the $\gamma$-network, referred to as the \textit{key view}. To match training conditions, this view should resemble the depth maps seen during training, as illustrated in ~\cref{fig:similarity}. We automatically choose the view with the largest object segmentation, crop to the masked region, and compute depth with Depth Anything V2~\cite{Yang24c}. We also train $\Phi$ on those predicted depth maps instead of synthetic ground-truth depth to further reduce the domain gap, as shown in our experiments.

\subsection{Estimating the Total Volume $\mathcal{V}$}
\label{sec:volume_estimation}

After estimating the occupancy ratio $\gamma$ as described above, we still require the total stack volume $\mathcal{V}$ to derive the object count from Eq.~\ref{eq:counting_eq}, and the unit volume $V_o$ if it is unknown. If cameras are uncalibrated, we run COLMAP~\cite{Schoenberger16a} to recover their poses and rescale them with a real-world reference. In industrial scenarios with fixed camera setups known {\it a priori}, simpler methods could be used.

Volume estimation $\mathcal{V}_{est}$ from multiple images is well understood. In our implementation, we begin with segmented images and retain only the container and objects by adding the mask as an alpha channel. Next, we optimize 3D Gaussian Splats~\cite{Kerbl23} from these images so the reconstruction includes only the container and objects. To measure the volume enclosed by the splats, we adapt the voxel carving algorithm. We initialize a voxel grid from the gaussian bounding box. Given the masks produced earlier and depth maps rendered from 3DGS, a voxel is removed if its projection in any view lies outside the mask or if its projected depth is smaller than the reconstructed depth. This procedure recovers objects and their containers, including partially filled boxes.

To remove the container from the reconstruction, we estimate its thickness $t$ and erode the voxels on all sides except the top by $t$, thereby refining the estimated volume to represent only the contents. The value of $t$ is predicted by an additional decoder $\Psi$ that takes the same encoded image features as our previous network $\Phi$ in~\cref{sec:occupancy}. To keep this prediction scale independent and predictable from 2D images alone, we express thickness as a ratio of the container's size. We supervise $\Psi$ with ground-truth thickness from our dataset and at inference we average the outputs of $\Psi$ over all images.
Combined, these algorithms yield a good estimate of the volume $\mathcal{V}$ spanned by the stacked objects.

\parag{Unit volume $v$.}
For the majority of industrial inspection settings, the unit volume $v$ of Eq.~\ref{eq:counting_eq} is known exactly because the object has been manufactured to a precise specification, $v$ can be computed from the simple geometry of the objects, or $v$ can be obtained from existing reference data, particularly for food items. When the unit volume of an object $v$ is not readily available, we estimate its value using the method described above from a set of images of a template object. This task is made easier by the absence of a container, and this is shape but not scene specific. For a new shape, the object is selected in a single frame and we can then use SAM2~\cite{Ravi24a} to generate a segmentation on all frames at once. The unit volume $v$ computed this way can then be used across all scenes containing this object.

\section{Experiments}
\label{sec:experiments}

\begin{figure}[t]
    \centering
    \captionsetup[subfigure]{justification=centering}
    \begin{subfigure}{0.24\columnwidth}
        \centering
        \includegraphics[width=\linewidth]{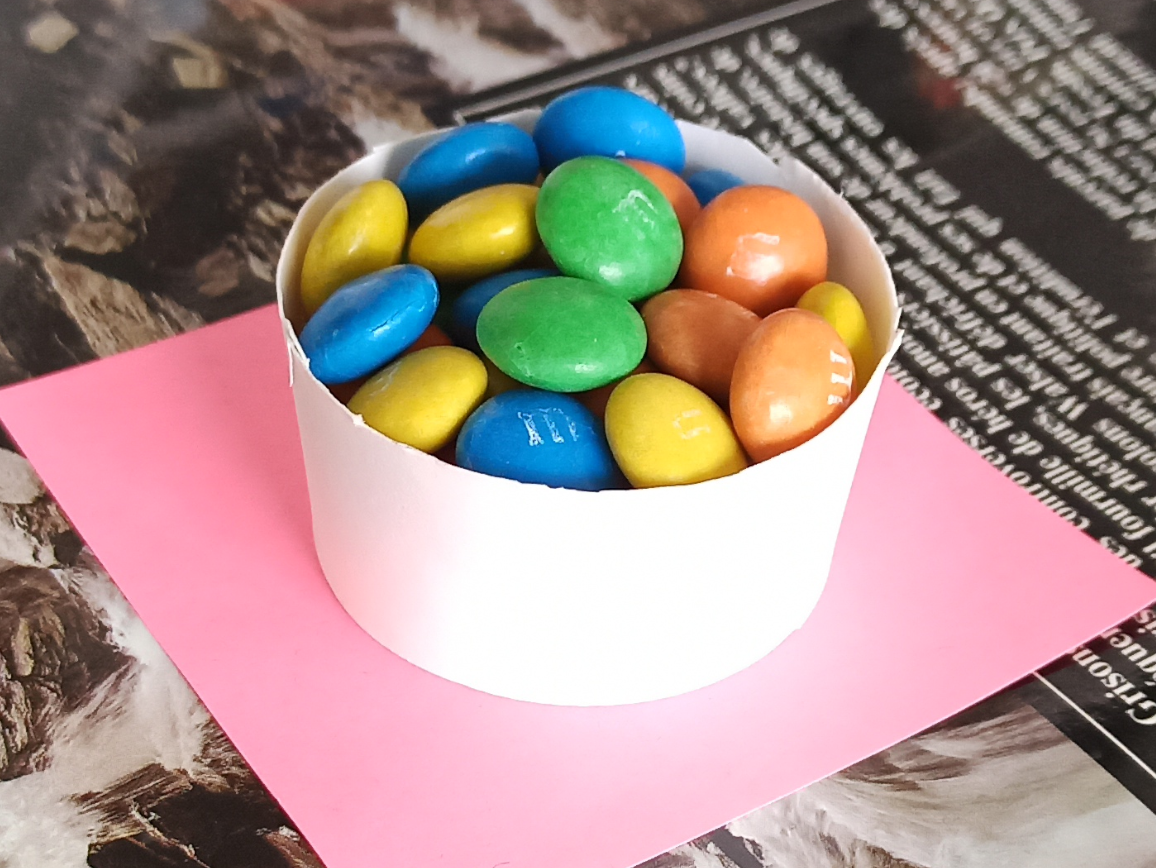} 
        \caption{$\mathcal{N}_{est} = 38$, \\$\mathcal{N}_{gt} = 36$}
    \end{subfigure}
    \hfill
    \begin{subfigure}{0.24\columnwidth}
        \centering
        \includegraphics[width=\linewidth]{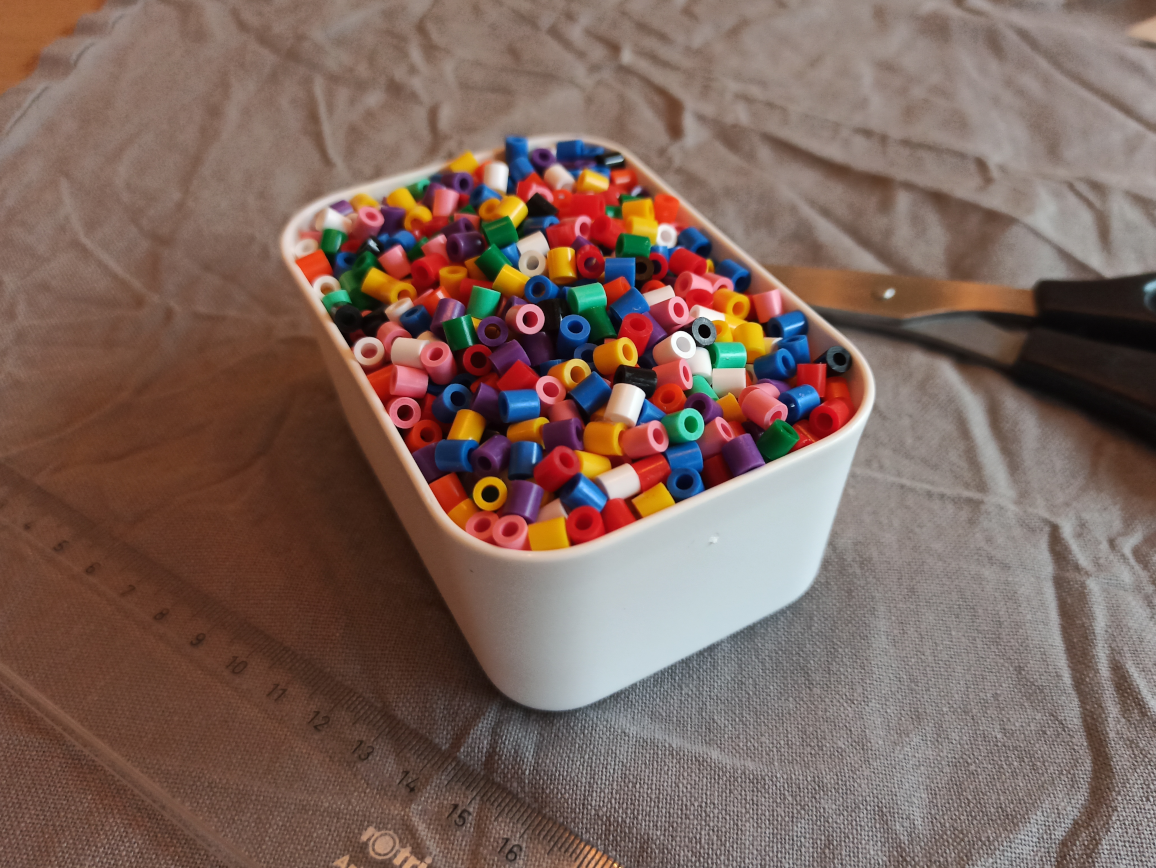} 
        \caption{$\mathcal{N}_{est} = 2133$, \\$\mathcal{N}_{gt} = 1830$}
    \end{subfigure}
    \hfill
    \begin{subfigure}{0.24\columnwidth}
        \centering
        \includegraphics[width=\linewidth]{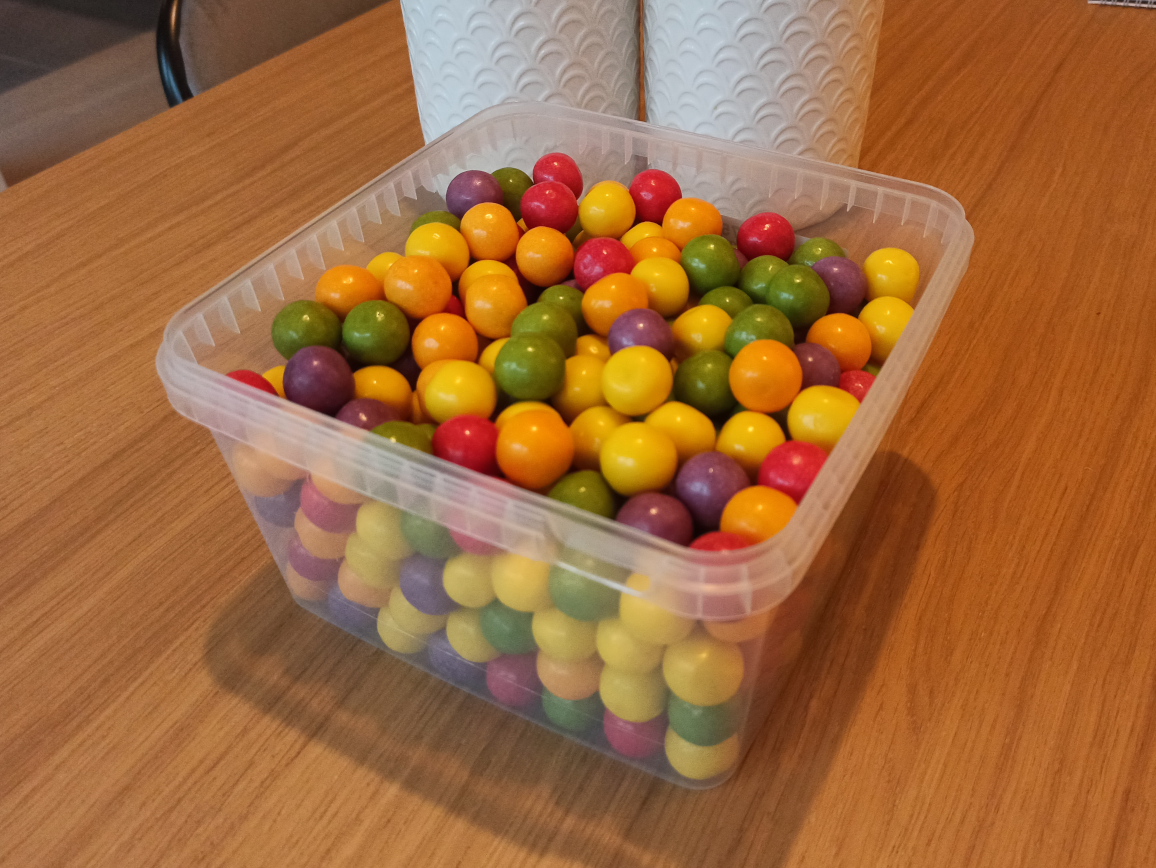} 
        \caption{$\mathcal{N}_{est} = 338$, \\$\mathcal{N}_{gt} = 397$}
    \end{subfigure}
    \hfill
    \begin{subfigure}{0.24\columnwidth}
        \centering
        \includegraphics[width=\linewidth]{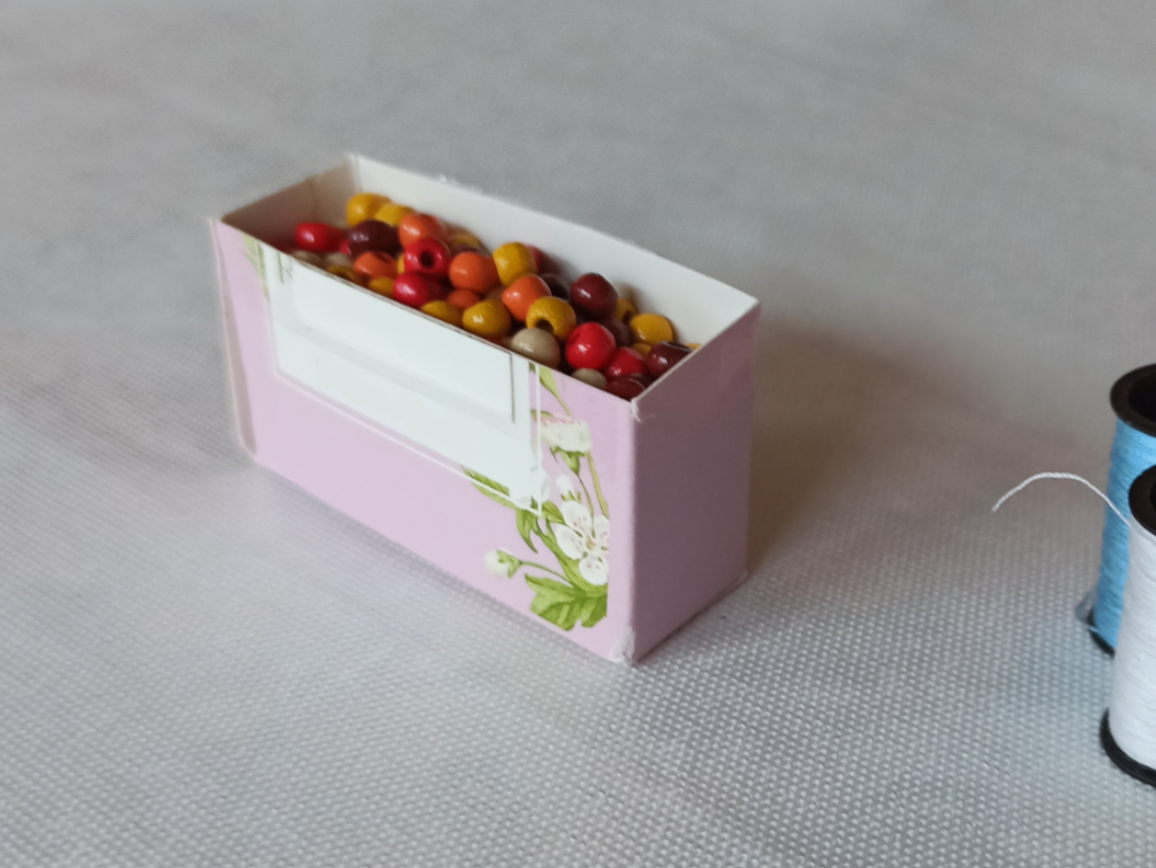} 
        \caption{$\mathcal{N}_{est} = 261$, \\$\mathcal{N}_{gt} = 300$}
    \end{subfigure}

    \caption{\textbf{Additional qualitative results.}}
    \label{fig:additional}
\end{figure}

\begin{figure*}[t]
    \centering
    \begin{subfigure}{0.32\textwidth}
        \centering
        \begin{subfigure}{0.48\textwidth}
            \centering
            \includegraphics[width=\linewidth]{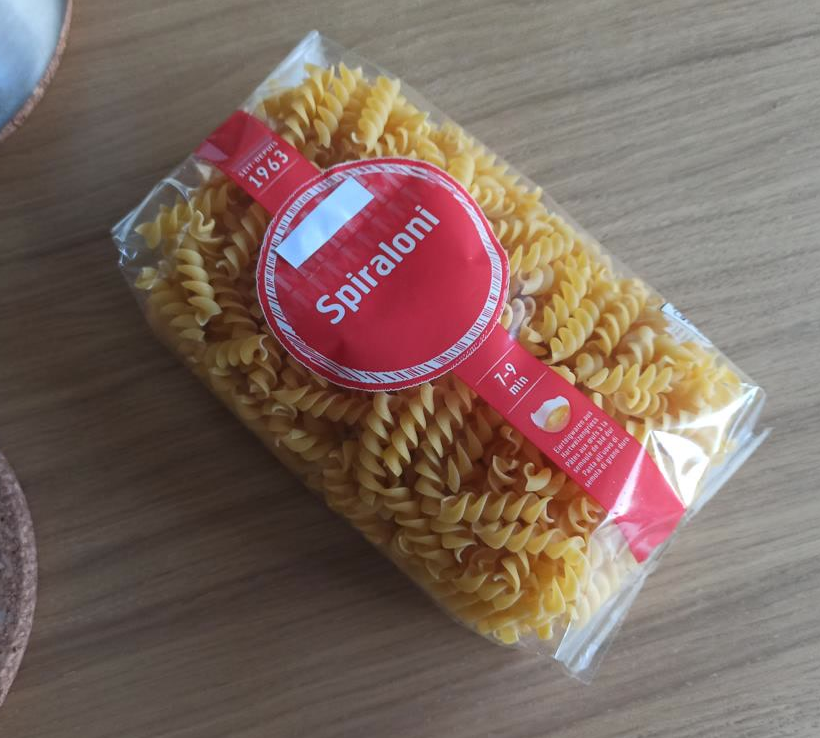} 
            \subcaption*{(a)}
        \end{subfigure}
        \hfill
        \begin{subfigure}{0.48\textwidth}
            \centering
            \includegraphics[width=\linewidth]{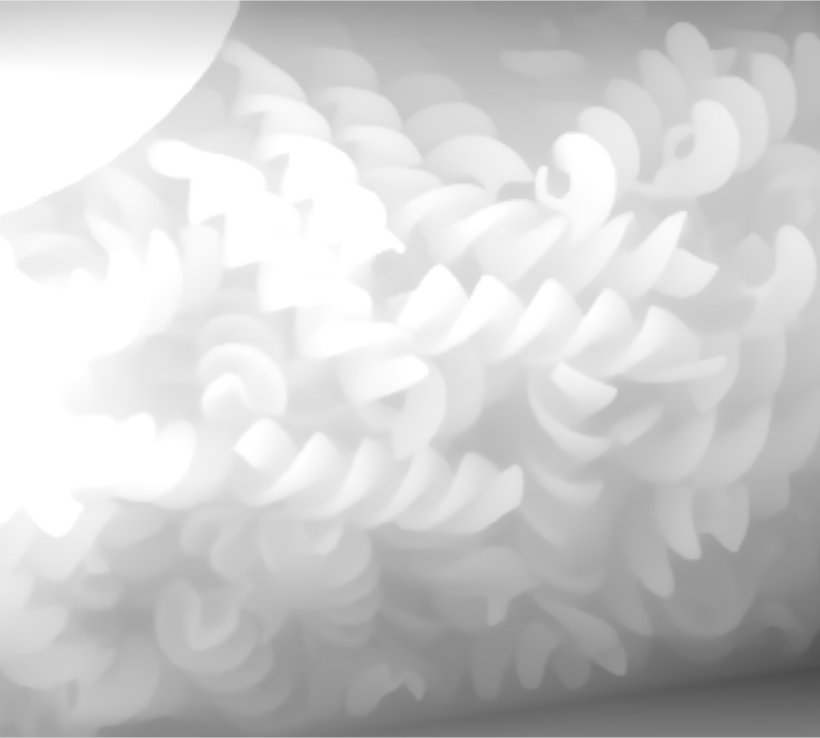}
            \subcaption*{(b)}
        \end{subfigure}

        \vspace{0.0em}

        \begin{subfigure}{0.48\textwidth}
            \centering
            \includegraphics[width=\linewidth]{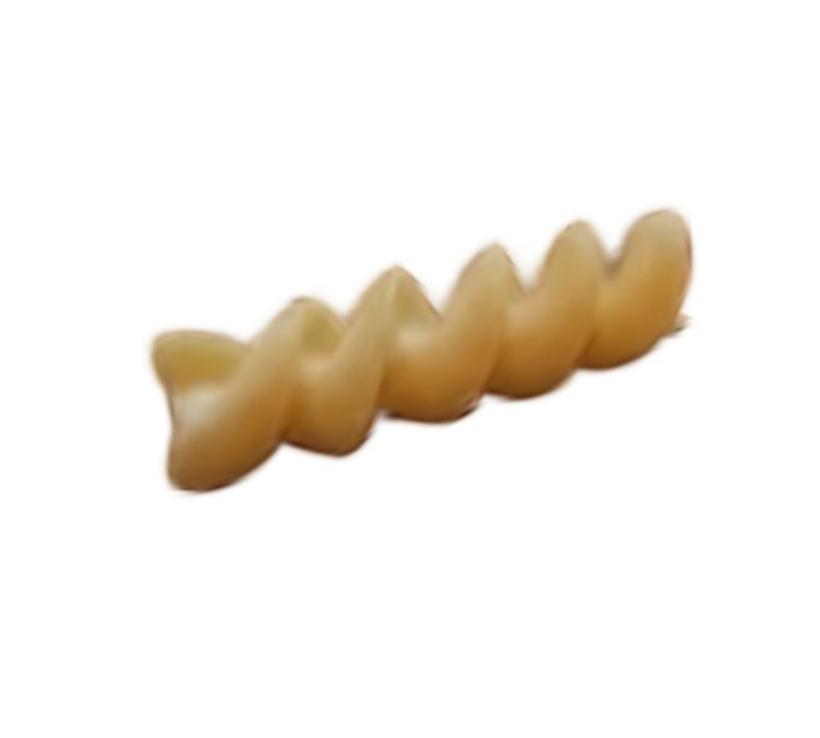}
            \subcaption*{(c)}
        \end{subfigure}
        \hfill
        \begin{subfigure}{0.48\textwidth}
            \centering
            \includegraphics[width=\linewidth]{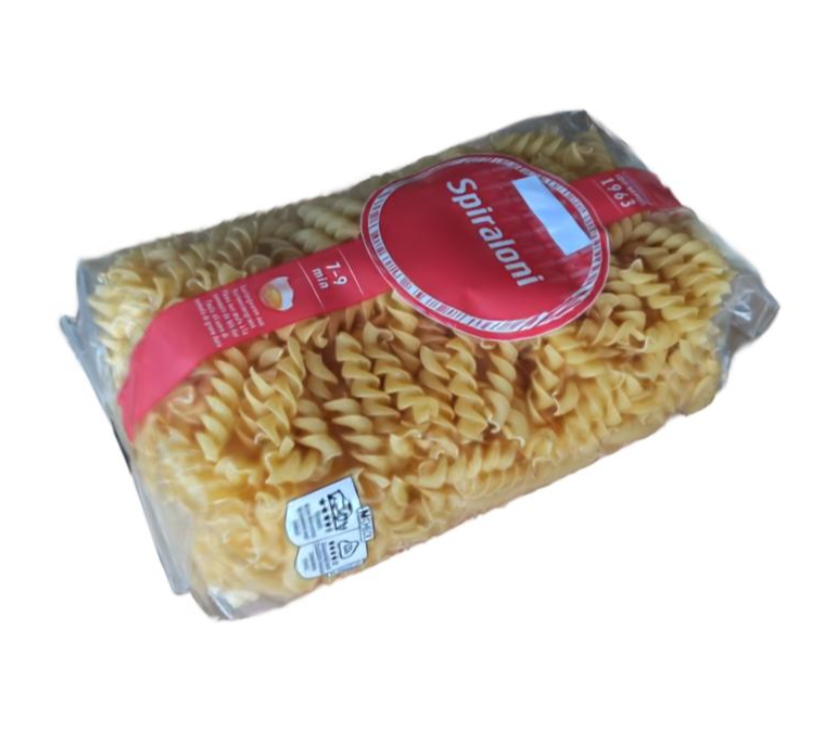}
            \subcaption*{(d)}
        \end{subfigure}
    \end{subfigure}
    \hfill
    \begin{subfigure}{0.32\textwidth}
        \centering
        \begin{subfigure}{0.48\textwidth}
            \centering
            \includegraphics[width=\linewidth]{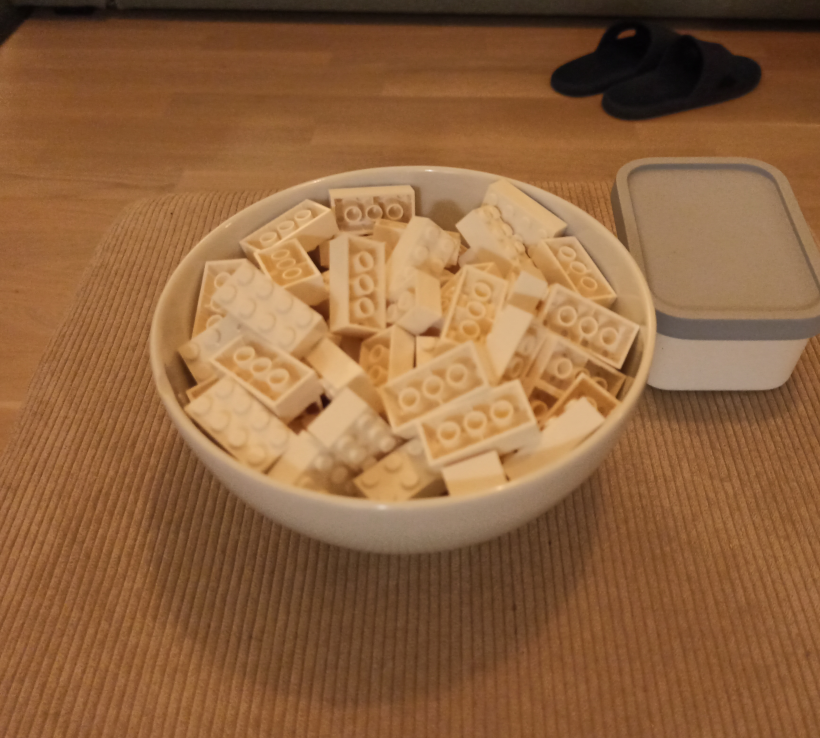}
            \subcaption*{(a)}
        \end{subfigure}
        \hfill
        \begin{subfigure}{0.48\textwidth}
            \centering
            \includegraphics[width=\linewidth]{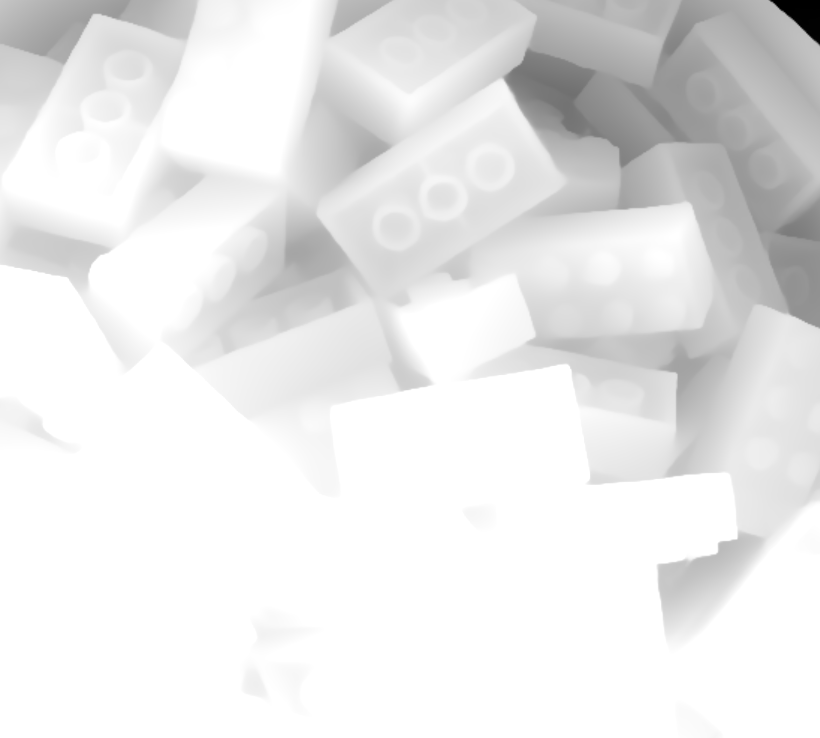}
            \subcaption*{(b)}
        \end{subfigure}

        \vspace{0.0em}

        \begin{subfigure}{0.48\textwidth}
            \centering
            \includegraphics[width=\linewidth]{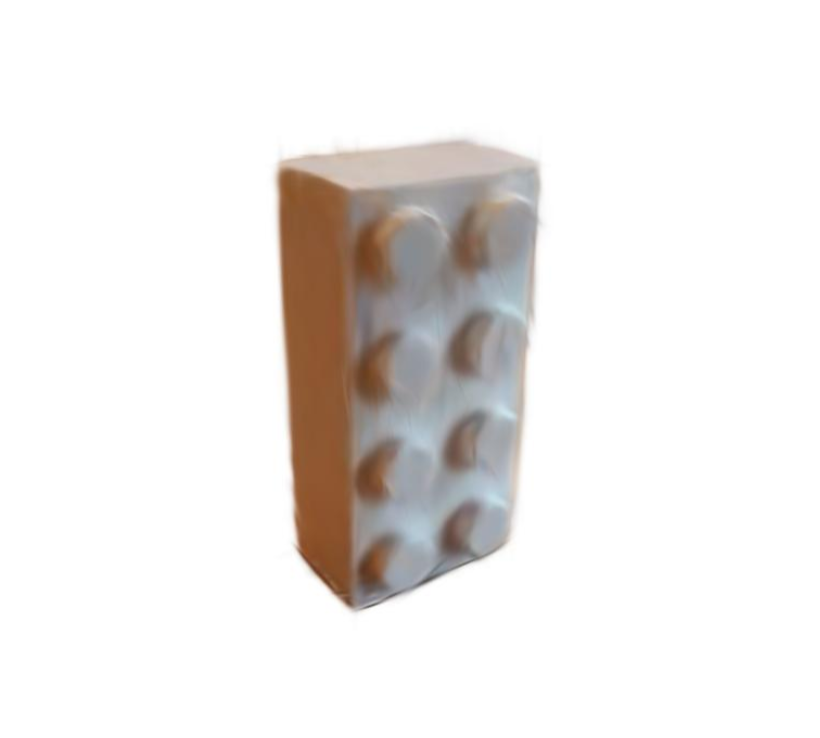}
            \subcaption*{(c)}
        \end{subfigure}
        \hfill
        \begin{subfigure}{0.48\textwidth}
            \centering
            \includegraphics[width=\linewidth]{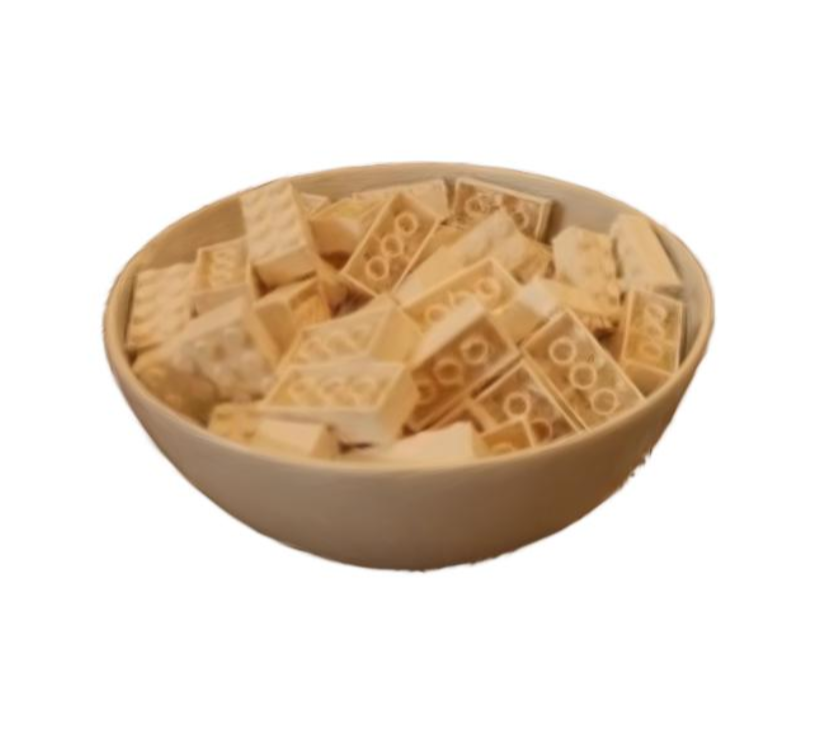}
            \subcaption*{(d)}
        \end{subfigure}
    \end{subfigure}
    \hfill
    \begin{subfigure}{0.32\textwidth}
        \centering
        \begin{subfigure}{0.48\textwidth}
            \centering
            \includegraphics[width=\linewidth]{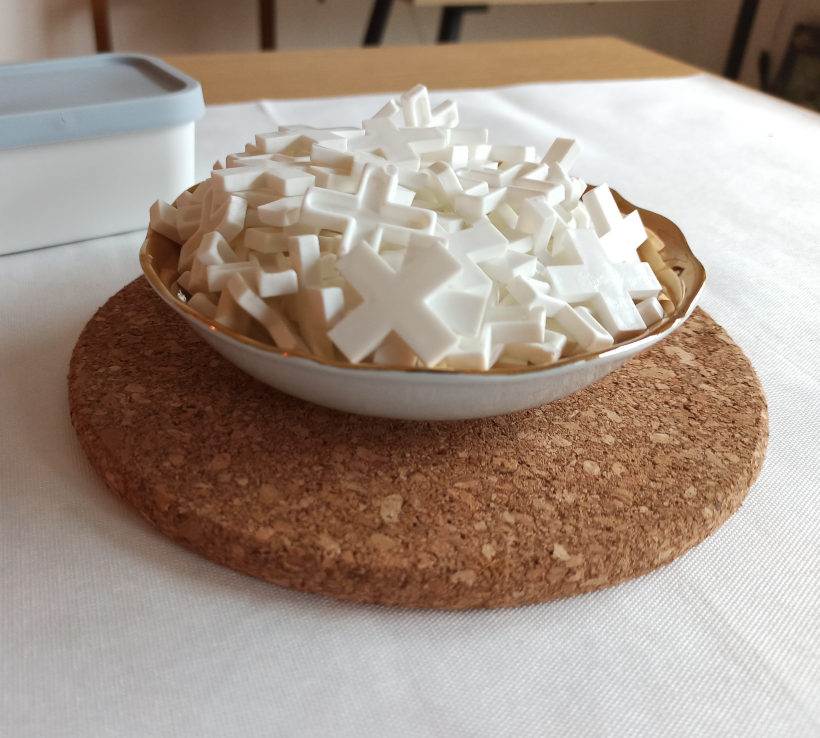}
            \subcaption*{(a)}
        \end{subfigure}
        \hfill
        \begin{subfigure}{0.48\textwidth}
            \centering
            \includegraphics[width=\linewidth]{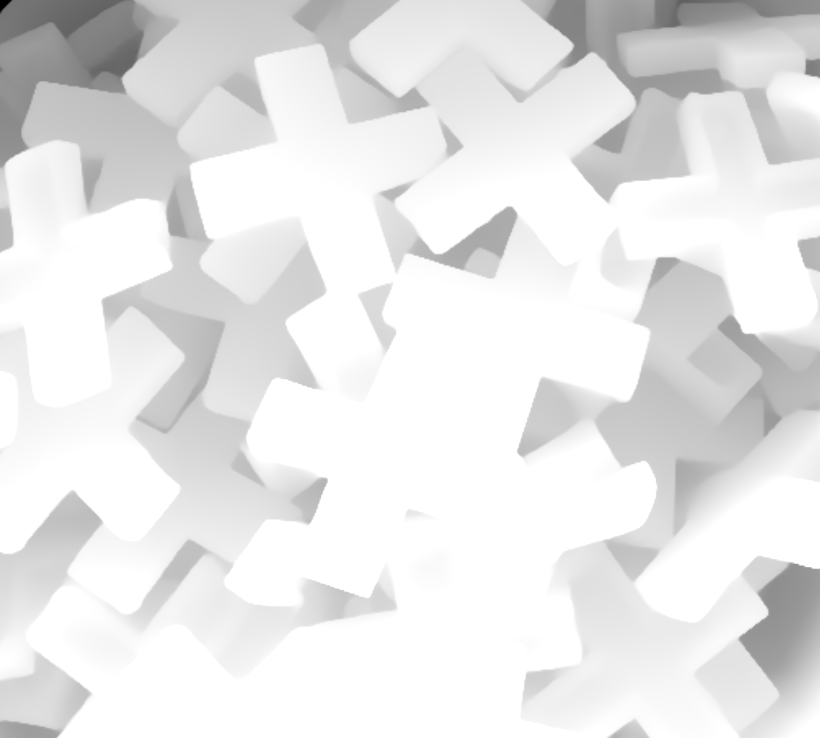}
            \subcaption*{(b)}
        \end{subfigure}

        \vspace{0.0em}

        \begin{subfigure}{0.48\textwidth}
            \centering
            \includegraphics[width=\linewidth]{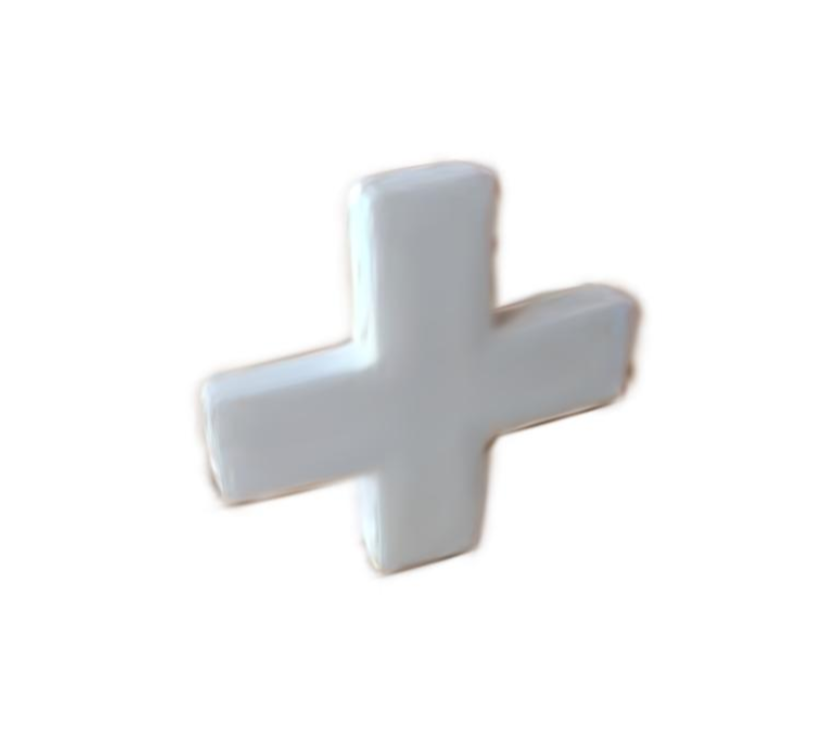}
            \subcaption*{(c)}
        \end{subfigure}
        \hfill
        \begin{subfigure}{0.48\textwidth}
            \centering
            \includegraphics[width=\linewidth]{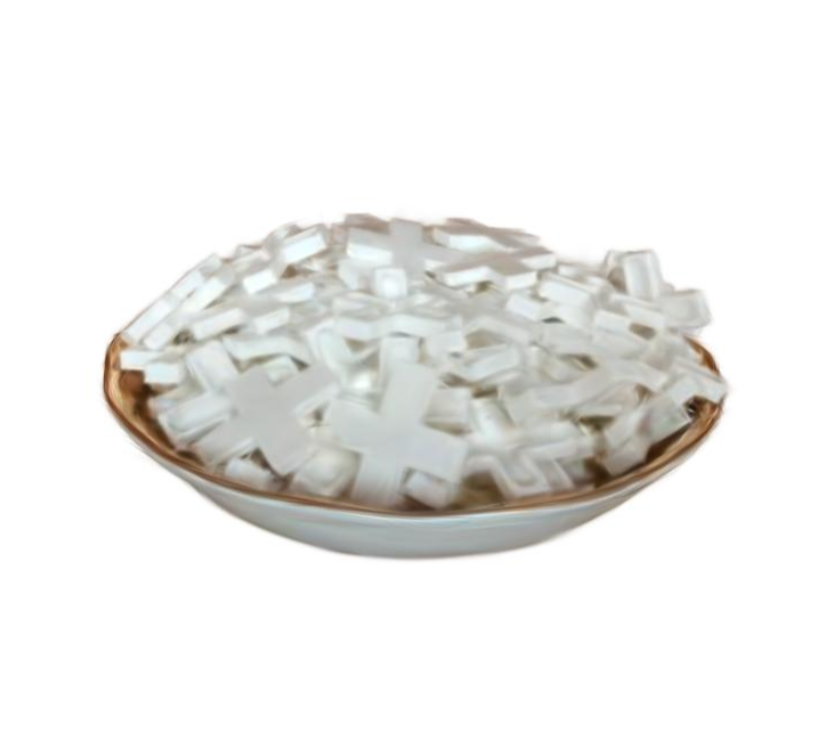}
            \subcaption*{(d)}
        \end{subfigure}
    \end{subfigure}

    \caption{\textbf{Intermediate results.} {From images (a), we find the key viewpoint and compute its depth (b) to estimate the occupancy ratio $\gamma$. Using the unit shape previously reconstructed from images of a template (c), and the overall reconstruction (d), we deduce the final count. (\textit{Pasta}: $\gamma_{est}=30.5\%$, $N_{est}=509$, $N_{gt}=588$ . \textit{Bricks}: $\gamma_{est}=31.8\%$, $N_{est}=73$, $N_{gt}=100$ . \textit{Crosses}: $\gamma_{est}=29.6\%$, $N_{est}=88$, $N_{gt}=116$ )}}
    \label{fig:intermediate_results}
\end{figure*}

We assess our method in three ways: measuring overall 3D counting accuracy in \cref{sec:eval_count}, occupancy ratio estimation in~\cref{sec:eval_gamma}, and volume estimation in~\cref{sec:eval_vol}. First, we use 100 scenes with physically simulated shapes from the ABC dataset~\cite{Koch19a}. Those scenes were isolated after generation and were not seen while training the occupancy ratio network. Another dataset consists of 2381 real images covering 45 scenes captured with a regular smartphone RGB camera without extra sensors. These captures provide multiple views around stacks of objects in containers, lying flat on a table, or still enclosed in packaging. 
Ground-truth counts are obtained manually for scenes below 500, or inferred from weight for larger counts. Finally, \cref{fig:intermediate_results} offers intermediate results to build intuition about our method's behavior, with more qualitative examples in \cref{fig:additional} and \cref{fig:teaser}. 

\begin{table}
    \centering
    \hspace{2em}
 \begin{small} 
    \begin{tabular}{l r r r r}
        \toprule
        & NAE $\scriptscriptstyle\downarrow$ & SRE $\scriptscriptstyle\downarrow$ & MAE $\scriptscriptstyle\downarrow$ & sMAPE $\scriptscriptstyle\downarrow$ \\
        \midrule
        BMNet+~\cite{Shi22a} & 0.91 & 0.87 & 320.50 & 158.87  \\
        SAM+CLIP~\cite{Kirillov23,Radford21} & 0.73 & 0.61 & 259.22 & 102.77 \\
        CNN & 0.66 & 0.48 & 235.74 & 98.44  \\
        ViT+H & 0.42 & 0.24 & 149.90 & 47.36  \\
        \hline
        Ours  & \textbf{0.22} & \textbf{0.09} & \textbf{79.48} & \textbf{27.65} \\
        \bottomrule
    \end{tabular}
    \vspace{-0mm}
    \caption{\textbf{Counting evaluation on 100 synthetic scenes.} 
    }
    \label{tab:eval_counting}
    \vspace{1em}

    \begin{tabular}{l r r r r}
        \toprule
        & NAE $\scriptscriptstyle\downarrow$ & SRE $\scriptscriptstyle\downarrow$ & MAE $\scriptscriptstyle\downarrow$ & sMAPE $\scriptscriptstyle\downarrow$ \\
        \midrule
        BMNet+~\cite{Shi22a} & 0.93 & 0.98 & 966.76 & 131.44 \\
        SAM+CLIP~\cite{Kirillov23,Radford21} & 0.94 & 0.99 & 980.33 & 124.31 \\ 
        CNN & 0.95 & 0.93 & 992.06 & 97.09  \\
        ViT+H & 0.94 & 0.93 & 979.29 & 91.45  \\
         \hline
        Human & 0.79 & 0.84 & 823.23 & 76.85 \\
        Human-Vote & 0.60 & 0.30 & 621.46 & 57.91 \\
        LlamaV 3.2 & 1.00 & 1.00 & 1037.5 & 190.48  \\
        \hline
        Ours (Color) & 0.57 & 0.27 & 607.98 & 74.33   \\
        Ours & \textbf{0.36} & \textbf{0.06} & \textbf{382.59} & \textbf{53.31} \\
        \bottomrule
    \end{tabular}
    \vspace{-0mm}
    \caption{\textbf{Counting evaluation on real-world scenes.}}
    \label{tab:eval_counting_real}
    \vspace{1em}

    \begin{tabular}{l c c c c}
        \toprule
        & MAE $\scriptscriptstyle\downarrow$ & RMSE $\scriptscriptstyle\downarrow$ & sMAPE $\scriptscriptstyle\downarrow$ & $R^2$ $\scriptscriptstyle\uparrow$\\ 
        \midrule
        \textit{DepthExtrapolated} & 0.36 & 0.38 & 77.43 & -6.04 \\
        \textit{DepthCorrected} &  0.10 & 0.12 & 34.80 & 0.28  \\
        Mean Estimator & 0.12 & 0.14 & 41.25 & 0.00 \\
         \hline
        Ours & \textbf{0.06} & \textbf{0.07} & \textbf{29.18} & \textbf{0.79} \\
        \bottomrule
    \end{tabular}
     \vspace{-0mm}
    \caption{\textbf{Occupancy ratio estimation.} We evaluate our method against three additional baselines that are tasked with prediction the occupancy ratio $\gamma$ from a depth map.}
    \label{tab:eval_gamma}  
     \end{small} 
\end{table}

\subsection{Metrics}

Several metrics are used to assess the accuracy of object counting and occupancy ratio estimation. Counts vary significantly across scenes, ranging from 19 to 20063. Consequently, the Mean Absolute Error (MAE), defined as $\text{MAE} = \frac{1}{n} \sum_{i=1}^n |y_i - \hat{y}_i|$ with \( y_i \) the ground truth count and \( \hat{y}_i \) the prediction, tends to emphasize scenes with high counts. To address this, we also report normalized metrics. We report the Normalized Absolute Error (NAE), Squared Relative Error (SRE), and Symmetric Mean Absolute Percentage Error (sMAPE), which scale errors relative to the ground truth. The NAE provides a measure of absolute error normalized by the total ground truth count across scenes, SRE highlights larger errors and penalizes significant deviations in high-count scenes, and sMAPE offers a normalized percentage error.

The exact formulas for the metrics used in~\cref{sec:experiments} are provided below. The NAE and SRE are defined as:
\begin{align*}
\text{NAE} &= \frac{\sum_{i=1}^n |y_i - \hat{y}_i|}{\sum_{i=1}^n y_i}, \hspace{1em}
\text{SRE} = \frac{\sum_{i=1}^n (y_i - \hat{y}_i)^2}{\sum_{i=1}^n y_i^2} \; .
\end{align*}
The Symmetric Mean Absolute Percentage Error (sMAPE) can be considered a normalized percentage error. It is expressed as follows:
\[
\text{sMAPE} = \frac{100\%}{n} \sum_{i=1}^n \frac{|y_i - \hat{y}_i|}{(y_i + \hat{y}_i) / 2},
\]
The formula of sMAPE ensures that errors are scaled symmetrically between the prediction and ground truth counts. Finally, the coefficient of determination, \( R^2 \), measures the proportion of variance in the ground-truth occupancy ratio $\gamma$ explained by our predictions
\[
R^2 = 1 - \frac{\sum_{i=1}^n (y_i - \hat{y}_i)^2}{\sum_{i=1}^n (y_i - \overline{y})^2},
\]
where \( \overline{y} \) is the mean of the ground truth counts. High values of \( R^2 \) indicate strong agreement between predictions and ground-truth values.

\subsection{Counting Evaluation}
\label{sec:eval_count}

To the best of our knowledge, there is no previous work on counting from multiple images that does not either assume all objects are visible or require additional sensors such as LiDARs. Therefore, we compare \acron{} against BMNet+~\cite{Shi22a}. This method predicts a density map over all pixels of an image and the estimated count is then inferred by summing over all pixels. We also compare against a combination of SAM~\cite{Kirillov23} and CLIP~\cite{Radford21}, where SAM generates many masks from an input image and CLIP uses negative and positive text prompts to identify masks for the object of interest. Finally, the total count is taken as the number of these masks. We compare these two baselines against our method in Tables~\ref{tab:eval_counting} and~\ref{tab:eval_counting_real}.

Early in our study, we attempted to directly predict object count from images. For this, we trained networks coined \textit{ViT+H} and \textit{CNN}, and we report their results in Tables~\ref{tab:eval_counting} and~\ref{tab:eval_counting_real}. These models perform poorly, especially on the real-world dataset, which prompted us to decompose the problem into occupancy and volume estimation. 

Next, we evaluated how well humans perform on this counting task. To do so, we organized a contest and asked participants to guess counts on the 45 real scenes. The contest registered 1485 guesses from 33 participants. We thus define the \textit{Human} baseline as the average of participant error metrics, and \textit{Human-Vote} as the error of the average guess across participants. Note that this second baseline should be stronger as participant errors tend to cancel out, which we observe in Tab.~\ref{tab:eval_counting_real}. However, results remain much worse than our approach. Interestingly, participants who spent more time did not perform better than their peers, highlighting the difficulty of this task. When asked about their method, most participants reported counting objects along each axis and multiplying, which proved ineffective. 

We also experimented with \textit{LlamaVision 3.2 11B}, a Large Language and Vision Model, on our counting task. While it described object appearance and scene composition well, its count estimates were completely off as shown in Tab.~\ref{tab:eval_counting_real}.

Overall, our approach beats all baselines, providing the first method to estimate stack counts with reasonable accuracy. Still, we note that performance remains better on synthetic data than on our real benchmark, likely due to the greater complexity of real scenes that often contain thousands of objects and are harder for volume estimation.

\subsection{Occupancy Ratio Evaluation}\label{sec:eval_gamma}

In this section, we now focus on evaluation of the occupancy ratio network alone. Since there is little work on occupancy ratio estimation, we implemented additional baselines to gain insights into our approach. 

Because we assume that the depth map contains enough information to predict the occupancy ratio, we define a first baseline we call \textit{DepthExtrapolated}. From the top view of the container, we compute the maximal depth using a monocular depth estimator and use it to normalize the depth map. Finally, we average the resulting values of the $K$ pixels, yielding the volume fraction estimate
$$
\gamma_{est}^{norm} = \frac{1}{K} \sum \frac{d_i}{d_{max}} \; .
$$
However, this first baseline tends to predict values lower than expected. Therefore, we defined a second one we dubbed \textit{DepthCorrected}, which uses linear regression to correct $\gamma_{est}^{norm}$ into a new estimate $\gamma_{est}^{corrected}$. This models the observation that depth maps with high variance tend to correspond to low occupancy ratios. We also compare with the mean estimator that predicts a mean percentage of $32.3\%$ occupancy ratio for all inputs.

As shown in \cref{tab:eval_gamma}, our method outperforms these three baselines by a significant margin. \textit{DepthCorrected} is better than the other two baselines, showing that depth information is indeed useful for this task. However, it still does not fully predict the occupancy ratio. We see this as evidence that the ratio $\frac{d_i}{d_{max}}$ alone is insufficient to predict $\gamma$, and our network learns to extract more informative cues from depth maps. We hypothesize that our occupancy ratio network captures additional geometric information, such as the influence of concavities on final volume occupancy.

\subsection{Volume Evaluation}\label{sec:eval_vol}

Given the novelty of volume computation directly from 3D Gaussian Splatting (3DGS) representations, there are currently no established, publicly released methods available for a direct comparative evaluation. Thus, to quantify the quality of our volume estimation scheme, we compare it to two simple geometric baselines: the Convex Hull and the $\alpha$-Concave Hull. The Convex Hull computes the volume of the smallest convex set containing all 3DGS centers, which results in a significant volume overestimation by ignoring concavities and fine structure. The $\alpha$-Concave Hull attempts to mitigate this by generating a tighter, non-convex boundary, but its volume calculation is highly sensitive to the parameter $\alpha$ and struggles to accurately delineate complex, sharp-edged objects.

As shown in~\cref{tab:evaluation_volume}, our method significantly outperforms both baselines across all metrics. The large discrepancy in performance underscores the limitations of relying on simple boundary approximations. These results validate that our technique accurately captures the intricate geometry and boundary of the underlying 3D structure, offering a robust and precise solution for volume calculation directly from the 3DGS scene representation.

\begin{table}[h!]
    \centering
    \begin{tabular}{l c c c}
       \toprule
       & MAE$\downarrow$ & RMSE$\downarrow$ & sMAPE$\downarrow$ \\ 
       \midrule
       Convex hull & 0.1847 & 0.2121 & 23.58 \\ 
       $\alpha$-concave hull & 0.4134 & 0.6724 & 54.07 \\ 
       Ours & \textbf{0.0455} & \textbf{0.0690} & \textbf{9.33} \\ 
       \bottomrule
   \end{tabular}
   \vspace{0em}
   \caption{\textbf{Evaluation of volume estimation.} }
   \label{tab:evaluation_volume}
\end{table}

\subsection{Ablation Study}

\begin{table}
    \centering
    \begin{tabular}{l r r r r}
        \toprule
        & NAE $\downarrow$ & SRE $\downarrow$ & MAE $\downarrow$ & sMAPE $\downarrow$ \\
        \midrule
        ($\mathcal{T}-$, $\mathcal{D}-$) & \textbf{0.22} & \textbf{0.09} & \textbf{79.48} & \textbf{27.65}  \\
        ($\mathcal{T}+$, $\mathcal{D}-$) & 0.28 & 0.11 & 100.12 & 30.93  \\
        ($\mathcal{T}+$, $\mathcal{D}+$) & 0.31 & 0.12 & 111.04 & 35.92 \\
        \bottomrule
    \end{tabular}
    \vspace{0.0em}
    \caption{\textbf{Ablation study on 3D counting.} If ground-truth depth maps are used during training, it is indicated as $\mathcal{T}+$, and $\mathcal{T}-$ otherwise. Similarly, for evaluation purposes if ground-truth depth-maps are used during validation, we indicate it as $\mathcal{D}+$.
            }
    \label{tab:ablation_counting}
    \vspace{1em}
    \begin{tabular}{l c c c c}
        \toprule
        & MAE $\downarrow$ & RMSE $\downarrow$ & sMAPE $\downarrow$ & $R^2$ $\uparrow$\\ 
        \midrule
        ($\mathcal{T}-$, $\mathcal{D}-$) & \textbf{0.06} & \textbf{0.07} & \textbf{29.18} & \textbf{0.79} \\
        ($\mathcal{T}+$, $\mathcal{D}-$) & 0.08 & 0.11 & 32.01 & 0.52 \\
        ($\mathcal{T}+$, $\mathcal{D}+$) & 0.10 & 0.13 & 37.35 & 0.32 \\
        \bottomrule
    \end{tabular}
    \vspace{0em}
    \caption{\textbf{Ablation study on occupancy ratio estimation.} }
    \label{tab:ablation_volume}
\end{table}

We conducted additional experiments to evaluate how sensitive our approach is to the depth maps produced by the monocular depth estimator. 

Because our synthetic dataset has ground-truth depth maps for both training and validation images, we evaluate them as follows. Recall from \cref{sec:occupancy} that, at training time, we use Depth Anything V2~\cite{Yang24c} depth maps, a setting we call $\mathcal{T}-$ for training without ground truth. Alternatively, we can use ground-truth depth maps during training, a setting we denote $\mathcal{T}+$. Likewise, at inference time, we can use the estimated depth map, which is our default, or the ground-truth one, referred to as $\mathcal{D}-$ and $\mathcal{D}+$, respectively. Consequently, the standard configuration of our method is $\mathcal{T}-,\mathcal{D}-$, and the other combinations are used only for ablation.

Results are reported in \cref{tab:ablation_counting} for counting and in \cref{tab:ablation_volume} for occupancy ratio estimation. Eliminating ground-truth depth maps and relying only on estimated ones, which is our standard procedure, performs best. We hypothesize that slight smoothing in the produced depth maps may prevent the model from overfitting to specific shape features in perfect depth maps.

Although the network is trained on synthetic data, this observation further confirms the generalizability of \acron{} to real data, since in practical scenarios such as the real data of \cref{tab:eval_counting_real}, ground-truth depth maps are not available.

Lastly, we also report in~\cref{tab:eval_counting_real} an ablated method \textit{Ours (Color)} where the $\gamma$-network takes an RGB image as input instead of a depth map. While this method still outperforms humans, it suffers from a significant performance drop and justifies the use of depth maps in our final approach.

\subsection{Validating the Counting Equation}\label{sec:counting_approximation_eval}

The counting equation \cref{eq:counting_eq} relies on minor assumptions detailed in \cref{sec:formula}, namely that the value of $gamma$ is approximately constant within the container and accurately measured in training samples. We validate these assumptions empirically using our synthetic dataset, which provides the ideal conditions necessary to isolate the formula's performance from external estimation errors.

To perform this validation, we utilize ground-truth parameters derived directly from the simulation. The individual object volume $v$ is computed from the mesh, $\gamma$ is measured at the end of physical simulation, and the total volume $\mathcal{V}$ is measured as the convex hull of the objects in the stack. Since these inputs are precisely known, any observed deviation stems solely from the inherent approximations of the counting equation itself or the definitions of $\gamma$ and $\mathcal{V}$.

\begin{figure*}[h]
    \centering
    \includegraphics[width=0.5\linewidth]{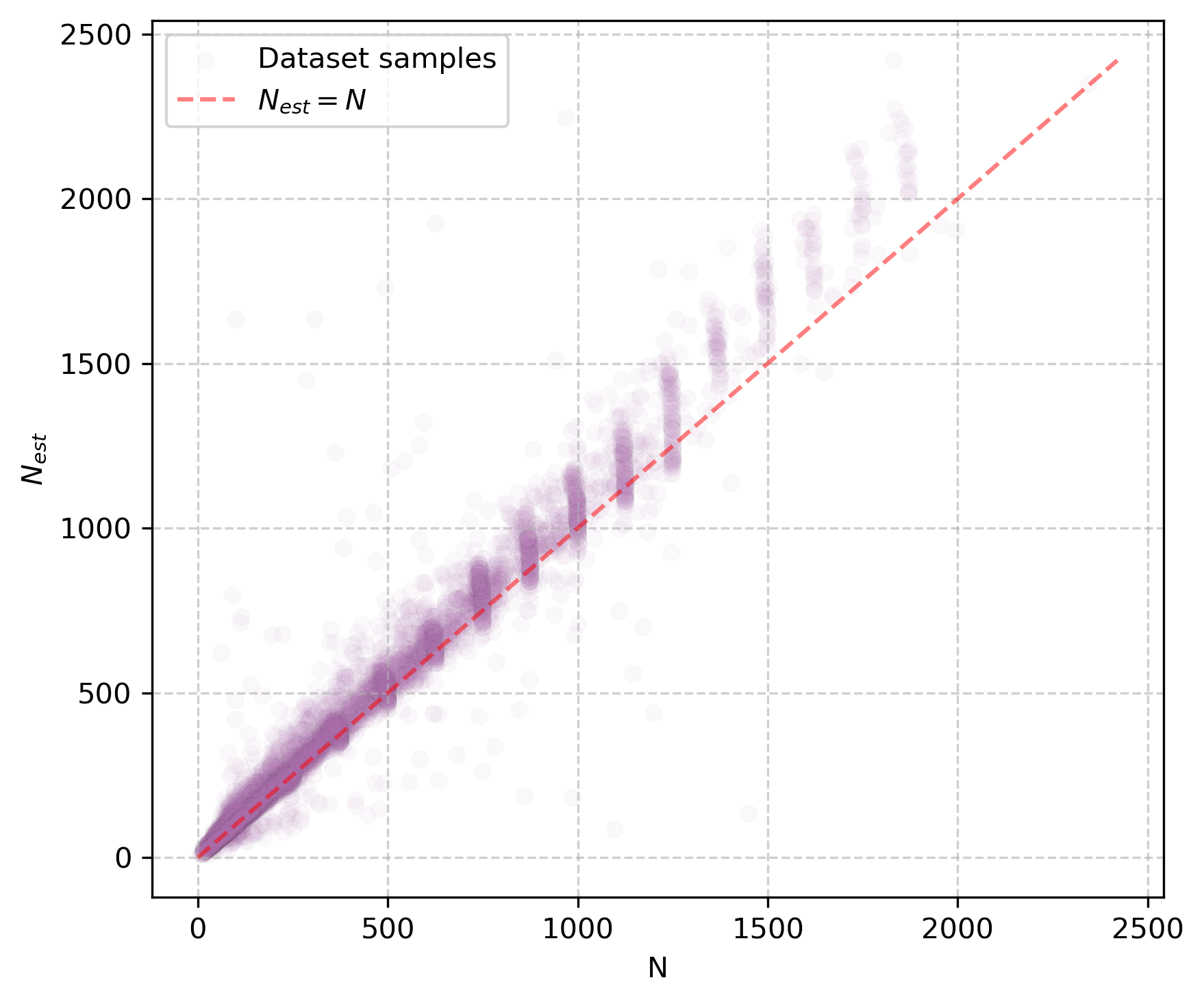}
    \caption{\textbf{Counting approximation.} In an ideal setting where $\gamma$, $v$ and $V$ are know, comparing the estimated counts with \Cref{eq:counting_eq} and ground-truth values demonstrates that this equation is a good approximation of the real count of objects in our released dataset, with a slight tendency to overestimate at larger counts.}
    \label{fig:counting_approximation}
\end{figure*}

\Cref{fig:counting_approximation} confirms the robust performance of the proposed formula. The comparison between the estimated count $\mathcal{N}_{est}$ and the ground-truth count $\mathcal{N}$ yields a Coefficient of Determination ($R^2$) of 0.9328. We observe a Mean Absolute Error of 49.21, which is relatively low given the scale of the stacks. While the correlation is strong, a slight tendency to overestimate is visible at larger counts. This is likely due to the convex hull volume $\mathcal{V}$ capturing small amounts of interstitial empty space at the stack boundaries, as well as minor local fluctuations in $\gamma$. Nevertheless, these results justify the use of our counting equation as a reliable proxy for object count, offering a more generalizable alternative to direct count estimation methods, which often struggle with high-density scenarios as demonstrated in \cref{sec:eval_count}.

\subsection{Additional Discussion on Border Effects}\label{sec:border_effects}

\begin{figure*}[t]
    \centering
    \includegraphics[width=\linewidth]{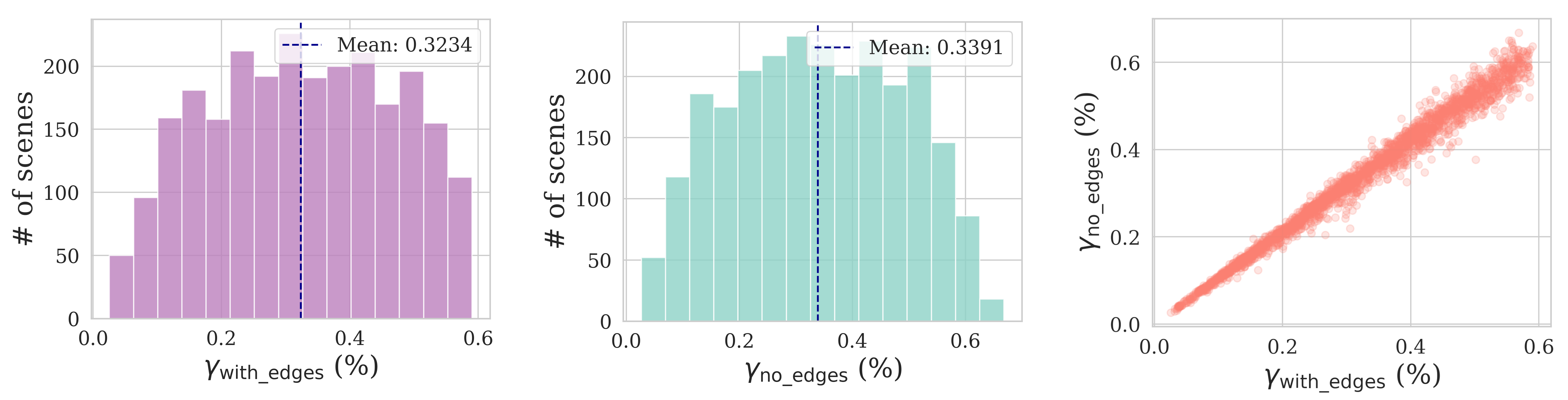}
    \caption{\textbf{Border effects.} Measuring the ground-truth $\gamma$ over the complete box or over only a smaller section makes little difference, indicating that our assumption of uniform $\gamma$ across the volume is justified.}
    \label{fig:supp_gamma_no_edges}
\end{figure*}

Our approach assumes that the occupancy ratio, \(\gamma\), is approximately uniform throughout the container. Generally, this assumption holds as the number of stacked objects increases. Yet, it neglects the influence of container borders, where objects tend to occupy less volume due to the boundary.

Quantitatively evaluating the impact of border effects and verifying the validity of our uniform \(\gamma\) assumption, we analyze the ground-truth volume ratio in two distinct ways using our large-scale synthetic dataset. Initially, we compute \(\gamma_{\text{with\_edges}}\) for the entire unit box, as described in the main paper. We also compute \(\gamma_{\text{no\_edges}}\) by measuring the volume ratio in a smaller sub-box of side length 0.5, centered within the unit box. The difference between \(\gamma_{\text{with\_edges}}\) and \(\gamma_{\text{no\_edges}}\) intuitively reflects the influence of border effects, allowing us to evaluate whether this assumption is justifiable for practical industrial applications.

Two histograms comparing the distributions of \(\gamma_{\text{with\_edges}}\) and \(\gamma_{\text{no\_edges}}\) are presented in \cref{fig:supp_gamma_no_edges}. Results indicate that both metrics follow highly similar distributions, with their mean values differing by less than 5\%. The mean value of \(\gamma_{\text{no\_edges}}\) is notably slightly higher than that of \(\gamma_{\text{with\_edges}}\), consistent with the intuition that density decreases near borders.

Further investigating the relationship between these two values, we provide a scatter plot of \(\gamma_{\text{with\_edges}}\) versus \(\gamma_{\text{no\_edges}}\) in \cref{fig:supp_gamma_no_edges}. This plot demonstrates a strong correlation between the two measures, particularly for objects with small volume ratios. Objects with high values of both \(\gamma_{\text{with\_edges}}\) and \(\gamma_{\text{no\_edges}}\) show minor discrepancies. We can attribute these differences to the relatively large size of these objects compared to the measurement box, which introduces noise in the estimation of \(\gamma\).

Collectively, these analyses confirm that \(\gamma_{\text{with\_edges}}\) and \(\gamma_{\text{no\_edges}}\) are highly consistent and can be used interchangeably without significant loss of accuracy. Our experiments rely on \(\gamma_{\text{with\_edges}}\) to train our occupancy ratio estimation network.

\subsection{Limitations and Future Work}

\begin{figure}[t]
    \centering
    \includegraphics[width=0.95\linewidth]{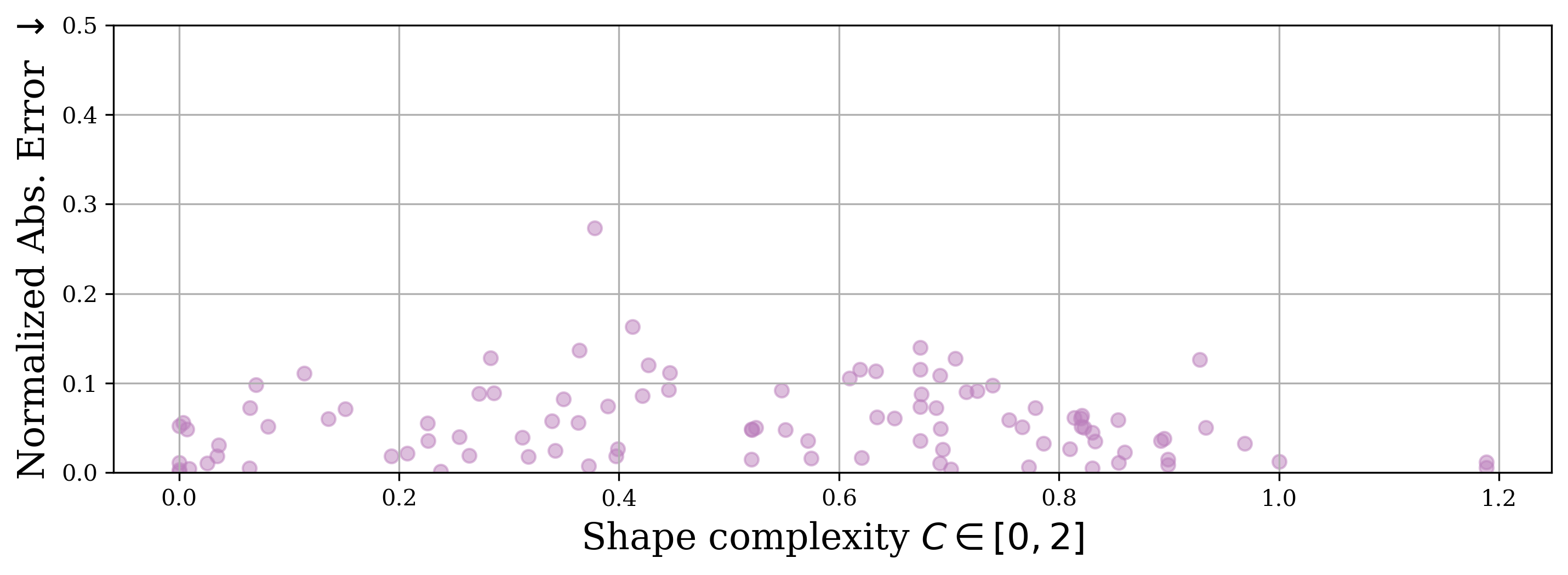}\vspace{-0.6em}
    \caption{\textbf{Complexity Analysis.} Each point represents the $\gamma$ occupancy ratio error for a shape in the validation set. 
    }\vspace{-0.6em}
    \label{fig:complexity}
\end{figure}
  
In order to evaluate the robustness of our method to complex shapes, we visualize in~\cref{fig:complexity} the error in occupancy ratio estimation as a function of shape complexity. Here, complexity is measured by summing a curvature term with the ratio of the shape's volume to the volume of its convex hull: 
$$
C = \frac{\kappa}{\|x_{max} - x_{min}\|_2 \kappa_0} + \frac{V_{hull}-V}{V_{hull}}
$$
where $\kappa$ is the integrated mean curvature of the shape, $\|x_{max} - x_{min}\|_2$ is a scaling factor and $\kappa_0$ is the maximum scaled $\kappa$ observed in the dataset. We observe only a slight error increase as shapes become more complex.

\begin{figure}[h!]
    \centering
    \begin{adjustbox}{max width=\columnwidth}
    \begin{tabular}{c}
        \begin{subfigure}[b]{0.32\linewidth}
            \centering
            \includegraphics[width=\linewidth]{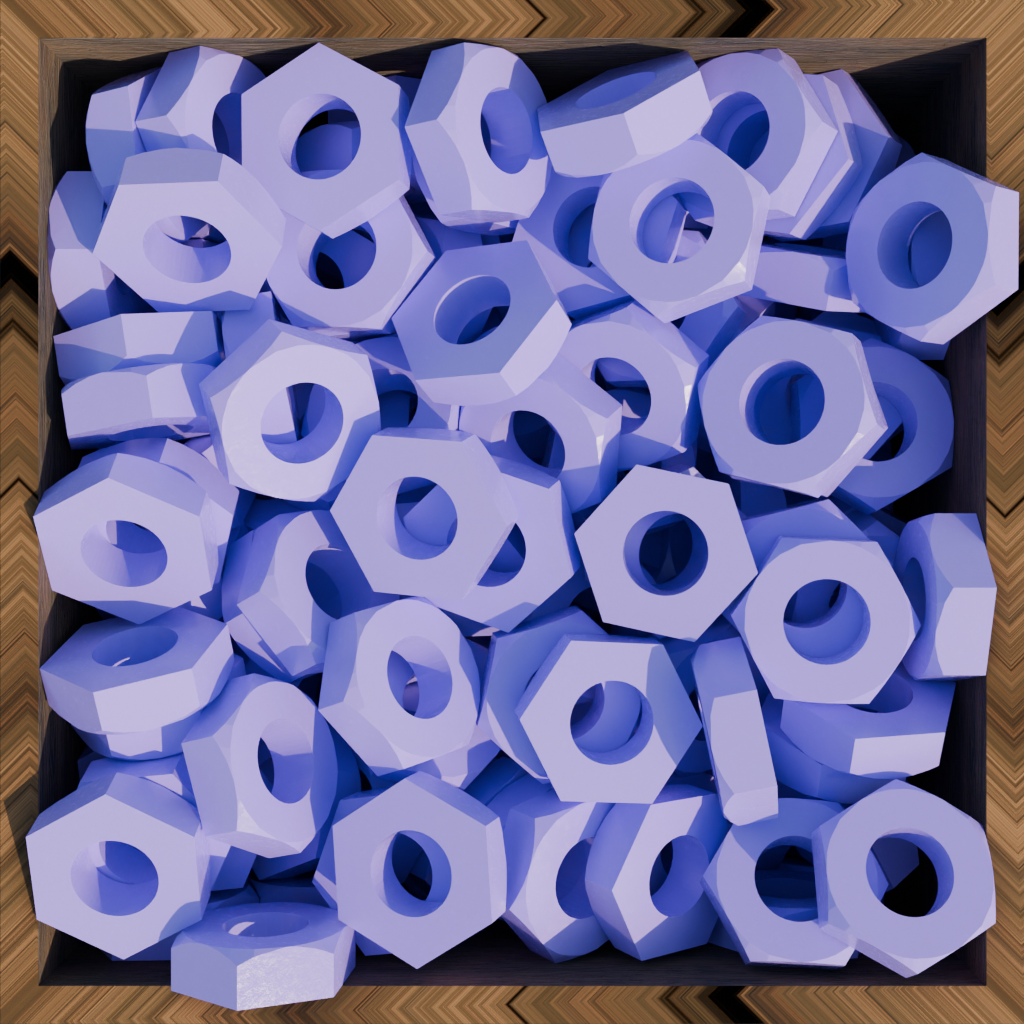}
            \caption{\scriptsize{Input image}}
        \end{subfigure}
        
        \begin{subfigure}[b]{0.32\linewidth}
            \centering
            \includegraphics[width=\linewidth]{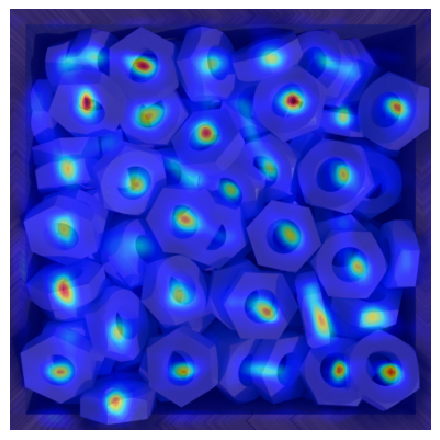}
            \caption{\scriptsize{BMNet+~\cite{Shi22a}}}
        \end{subfigure}
        
        \begin{subfigure}[b]{0.32\linewidth}
            \centering
            \includegraphics[width=\linewidth]{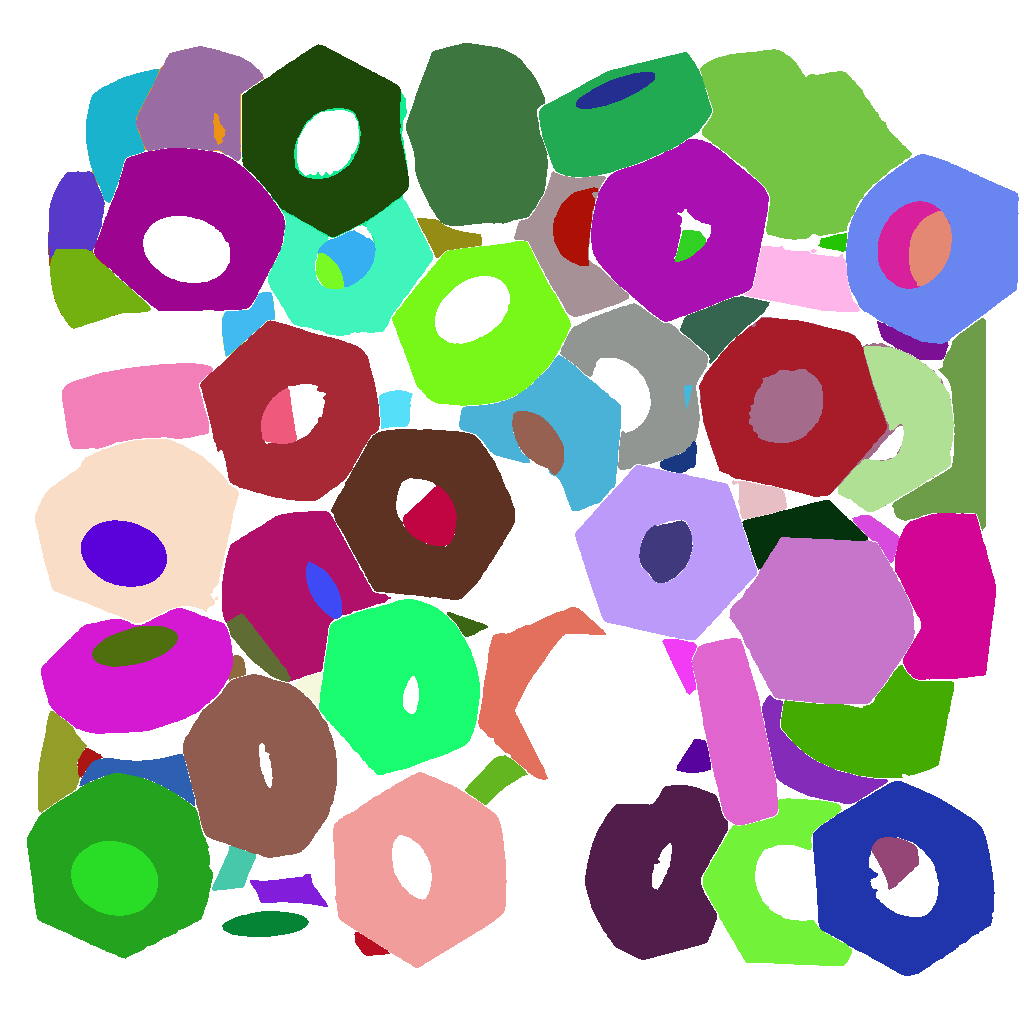}
            \caption{ \scriptsize{SAM\cite{Kirillov23}+CLIP\cite{Radford21}}}
        \end{subfigure}
    \end{tabular}
    \end{adjustbox}
    \vspace{-1em}
    \centering
    \caption{\textbf{Instance localization.} Previous methods also produce interpretable results, representing a promising direction for future work.}
    \label{fig:limitations}
\end{figure}

Many earlier methods attempt to localize the objects being counted, unlike ours, thus increasing the interpretability and usability of the results. However, these localizations are often erroneous when objects are stacked together, as illustrated in \cref{fig:limitations}, greatly limiting their applicability. Another possible direction for further enhancements lies in integrating a robust localization of visible instances and an estimation of a plausible configuration of invisible instances.

\section{Conclusion}
\label{sec:conclusion}

We presented a new method to count sets of stacked, nearly identical objects, which are particularly common in industrial inspection where objects are often placed in bulk in large containers. Occlusions and irregular arrangements make accurate counting difficult in this setting. By decomposing the counting task into complementary subproblems, separately estimating the 3D volume of the stacks and the fraction of that volume occupied by objects, we proposed an effective solution that is easy to implement and significantly outperforms humans on this challenging task. 

Our experiments indicate that performance can decline with increased geometric complexity or visually complex scenes. For future work, we will focus on training the volume occupancy estimator to overcome these challenges. 
Overall, we believe our method and proposed datasets will open new applications and encourage future efforts centered on stacks of 3D objects, including 3D reconstruction, counting, and 3D scene understanding. These new methods will have the potential to fundamentally reshape our current industrial inspection pipelines, enabling automated quality control, reducing manual labor costs, and improving accuracy in inventory management. 

\section{Declarations}

\subsection{Acknowledgements.}
We would like to sincerely thank all the participants who spent time to play our counting game and provided us with valuable data on human performance at this task. In particular, we would like to congratulate Martin Engilberge, who gave the most accurate estimates and has been awarded a box of chocolates. We would also like to thank Nicolas Talabot, Víctor Batlle and Adriano D'Alessandro for insightful discussions. 

\section{Competing interests.}
This work was supported in part by the Swiss National Science Foundation. There is no other competing interest involved in this work.

\bibliography{sn-bibliography}

\newpage
\begin{appendices}

\section{StackCounting-Real dataset.}
Our dataset contains 45 real scenes where the objects to count can be any stack of items that are at least partially visible. Examples include stacked objects on a table or on the floor, objects in containers such as bowls or boxes, or objects still in their packaging. Such configurations are common in industrial inspection settings, where products may be arranged in bins, boxes, or still within their original packaging.

\parag{Cameras.}
A standard RGB smartphone camera captures 30-60 pictures of the scene from various angles, forming a semisphere surrounding the objects and their container. Downscaling these images to approximately 600 pixels wide reduces memory usage and facilitates processing with COLMAP \cite{Schoenberger16a}. Measuring an arbitrary object within the scene allows us to scale the camera measurements and align the unit distance of the scene with a meter in the real world. 

Our initial experiments explored triangulation methods using two pairs of corresponding points across images. Calculating a 3D distance from these correspondences would enable scaling of the scene. Small inaccuracies in point matching, however, led to significant variations in the scaling, making this approach unstable. Reconstructing the 3D scene with 3DGS \cite{Kerbl23} instead allows us to measure the 3D distance directly within the reconstruction. Rescaling the scene to match the reference measurement becomes straightforward with this approach. The 3D point cloud generated by COLMAP \cite{Schoenberger16a}, which is used as an initialization by 3DGS, is also scaled accordingly.

\parag{Unit volume.}
Each scene requires knowledge of the unit volume of the object being counted. Complex shapes necessitate determining the unit volume $v$ by 3D reconstruction, similar to the method described in our main paper. Many common food items, such as kidney beans or corn, have this information readily available online. Other scenes allow volume approximation, for example in the case of the beads in \cref{fig:teaser}, by subtracting the volume of a cylinder from that of a sphere. In industrial inspection applications, unit volumes are often known from product specifications, though irregular or damaged items may require similar reconstruction-based approaches.

\parag{Pre and Post-processing.}
This method enables us to capture 45 scenes consisting of various items in different environments. Complexity varies across scenes, from simple quasi-spherical objects in containers to more challenging configurations, such as complicated shapes still in packaging (e.g., in the \textit{pasta} scene). Several items appear in multiple scenes, where the container and location are modified to create a new setting.

\section{StackCounting-Synthetic Dataset.} 
We use Blender, a free and open-source 3D creation suite that supports Python scripting, to generate our large-scale synthetic dataset. This allows us to implement a fully automated generation pipeline, which is mainly composed of two steps: simulation and rendering.

\parag{Simulation.}
Batches of objects, arranged in a $4 \times 4 \times 5$ grid, are dropped into a box positioned at $(0, 0, 0.5)$. The initial side length of the box is $1$ and its thickness is taken between $0.01$ and $0.05$, then a random scaling is applied on each dimension. After convergence of the simulation, checking whether the union of the objects intersects with an invisible cube placed directly on top of the box determines whether to continue. When an intersection occurs, the simulation stops, and objects outside the box are deleted. When no intersection is detected, a new batch of objects is added, and the simulation is performed again. In some random cases, simulation is interrupted to allow for partially full boxes.

The convex hull is used to compute collision between objects. Using the triangle mesh itself would be ideal, however this becomes far too costly when physically simulating thousands of shapes with tens of thousands of triangles. Experiments with using convex hulls first, and then refining with additional frames using the triangle mesh, revealed that this approach remains extremely costly and computationally very unstable, leading to objects being ejected outside the box due to the change in collision computation. 

\parag{Rendering.}
Our rendering employs textures randomly sampled from 3 possibilities for the box, five textures for the ground, and a random material for each model chosen from one of the following: a realistic grey metal texture, a red metallic texture, or a plastic material with a randomly selected color. 

The first view is always rendered directly above the box, looking downwards, which we call the \textit{nadir} view. The validation dataset also includes 29 additional views on the unit sphere, each observing the box from different angles. Blender's Cycles rendering engine performs the rendering. Ground-truth depth maps and masks are also generated, separating the ground, box, and objects in the images.

\parag{Pre and Post-processing.}
Pre-processing filters out unsuitable meshes, such as those with multiple connected components or excessive size, in addition to the simulation and rendering steps. Physical simulations can sometimes be unstable or fail, requiring removal of a small fraction of results in post-processing. These include cases where the unit volume is too small or where too few objects remain in the box in the final frame.

Exporting the calibrated camera parameters in a format compatible with nerfstudio \cite{Tancik23} completes the pipeline. These cameras are not generated by COLMAP, so they do not include a 3D point cloud that 3DGS can use as initialization. This poses a challenge, as a fully random initialization may generate distant Gaussian points outside the cameras' range, which are not removed and interfere with the volume estimation. To address this challenge, we instead generate a set of 100 grey points within the unit cube, centered at $(0, 0, 0.5)$. This initialization proves sufficient to quickly produce a faithful 3D reconstruction and resolves the aforementioned issue.

\section{Additional comparison distinguishing visible and invisible objects.}

\begin{figure}[t]
    \centering
    \hfill
    \begin{minipage}{0.45\columnwidth}
        \includegraphics[width=\linewidth]{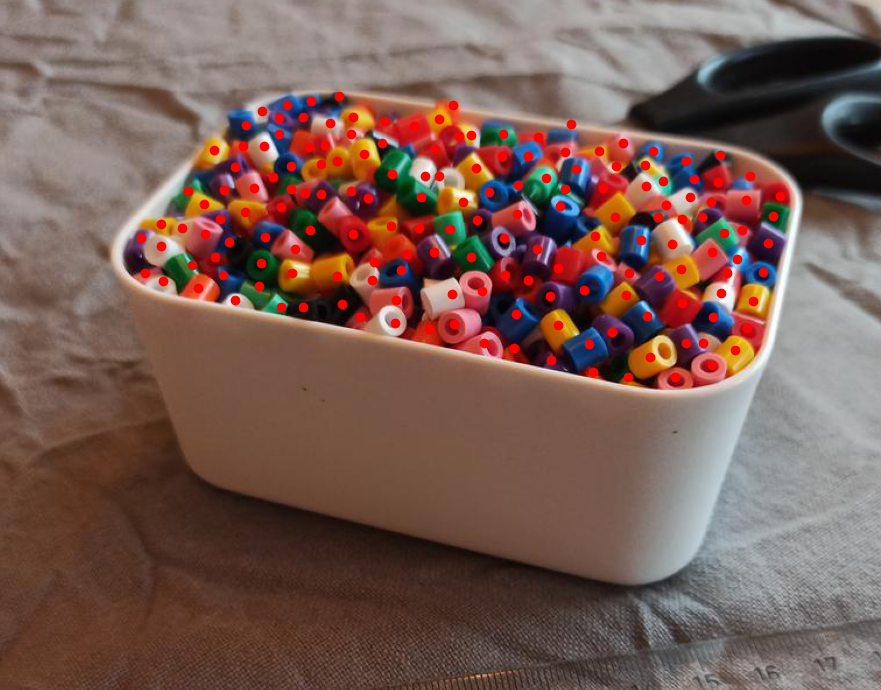}
    \end{minipage}
    \hfill
    \begin{minipage}{0.45\columnwidth}
        \includegraphics[width=\linewidth]{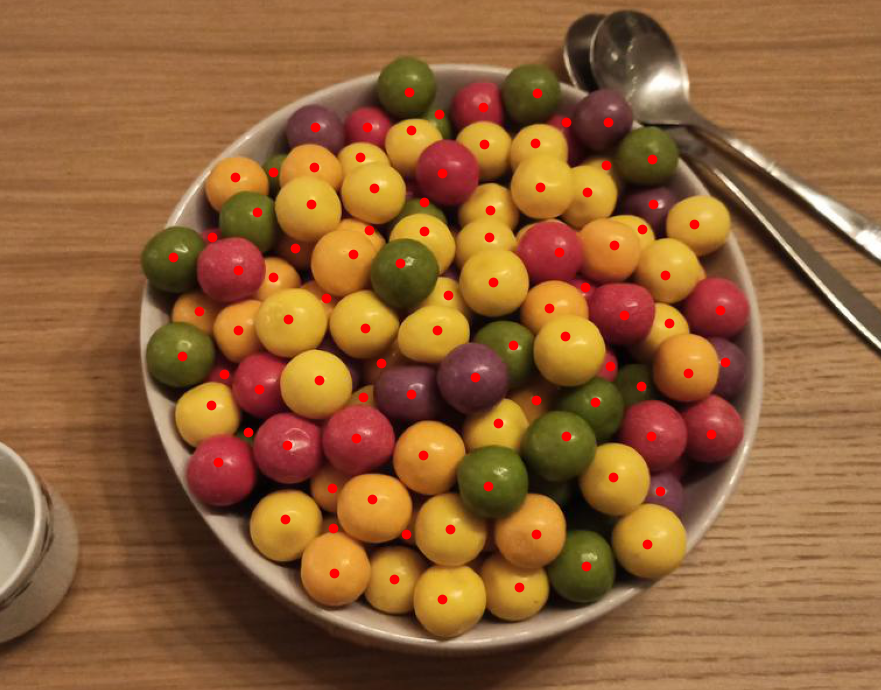}
    \end{minipage} 
    \hfill 
    \caption{\textbf{Localization of visible objects.} 
    Left: $\mathcal{N}_{vis}=168$, Right: $\mathcal{N}_{vis}=96$}
    \label{fig:localization}
\end{figure}

To the best of our knowledge, our method proposes the first solution for this task. Comparison methods in \ref{sec:experiments} are initially designed to count only visible objects, and perform poorly on our dataset. This section distinguishes between visible and invisible objects in stacks and evaluates baseline methods against both counts. Manual annotation of the locations of visible objects in real scenes enables this evaluation, and we provide the visible count and locations in our released datasets. \cref{fig:localization} displays two examples.

\cref{tab:eval_visible} presents the results of this experiment. Distinguishing similar objects clumped together remains difficult for 2D counting methods, even though their numbers improve. Performance remains much lower on these challenging scenes than on traditional 2D counting benchmarks as a result. Such limitations are particularly problematic in manufacturing and retail, where such scenes are particularly prevalent.

\begin{table}
    \centering
    \begin{small} 
    \begin{tabular}{l r r r}
        \toprule
        & NAE $\scriptscriptstyle\downarrow$ & SRE $\scriptscriptstyle\downarrow$ & sMAPE $\scriptscriptstyle\downarrow$ \\
        \midrule
        \textit{Visible objects only} \\
        BMNet+ [21] & \textbf{0.51} & \textbf{0.65} & 51.28 \\
        SAM+CLIP [7,16] & 0.57 & 0.81 & \textbf{50.84} \\
        \hline
        \textit{All objects} \\
        BMNet+ [21] & 0.93 & 0.98  & 131.44 \\
        SAM+CLIP [7,16] & 0.94 & 0.99  & 124.31 \\ 
        Ours & \textbf{0.36} & \textbf{0.06} & \textbf{53.31} \\
        \bottomrule
    \end{tabular}
    \caption{\textbf{Counting visible and invisible objects separately.}
    }
    \label{tab:eval_visible}
\end{small} 
\end{table}

\begin{figure*}[t]
    \centering
    \includegraphics[width=\linewidth]{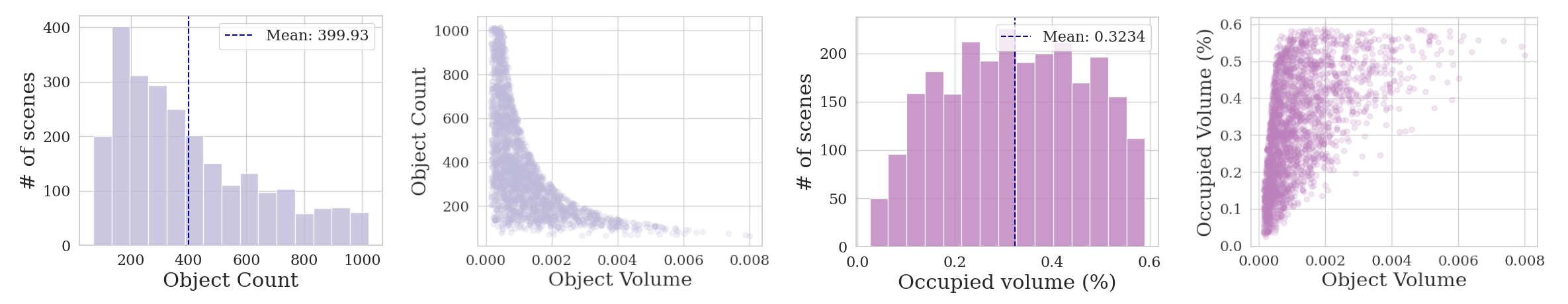}
    \caption{\textbf{Dataset statistics.} The histograms represent the distributions of object count and occupancy ratio, respectively, and each bar plots the number of scenes in a given bin.  In scatter plots, each point represents a physically simulated 3D scene. In particular, the occupancy ratio $\gamma$ spans a large range between $1\%$ and 65$\%$}
    \label{fig:dataset_stats}
\end{figure*}

\section{Implementation details}
\label{sec:implementation_details}

The nerfstudio library~\cite{Tancik23} is used for 3D reconstruction, specifically the \textit{splatfacto} method built on top of the gsplat library~\cite{Ye24a}. We thank the contributors of all the aforementioned 
libraries.

Pretrained models for depth estimation and mask generation are also used in our pipeline. The \textit{vitl} model from Depth Anything v2~\cite{Yang24c} is employed for depth estimation and the \textit{sam2.1\_hiera\_large} model from SAM2~\cite{Ravi24a} for mask generation. State-of-the-art models ensure high-quality and robust outputs across diverse scenes, which is crucial for reliable industrial inspection systems that must operate under varying lighting conditions and object arrangements.

Generating the dataset uses CPUs only, greatly reducing its production cost and environmental impact. Remaining operations are fairly light and performed locally on a 4080 Mobile GPU, taking up only a few gigabytes of VRAM and being completed in a couple minutes.

\section{Architecture details}
\label{sec:architecture_details}

A DinoV2~\cite{Oquab23} encoder model that produces pixel-aligned features is utilized in our architecture. DinoV2 downscales the input image by 14, so we feed it an image of size 448 x 448 to produce a 32 x 32 x 768 feature image. The pretrained weights of the \textit{dinov2\_vitb14} model are specifically used and frozen during all subsequent learning.

Predicting a scalar value from the \(32 \times 32 \times 768\) feature image produced by DinoV2, we employ a series of convolutional layers to progressively reduce both the spatial dimensions and the number of channels. Successively, the convolutional layers reduce the channel dimension from the initial 768 down to 512, 256, 128, and finally 64. The spatial dimensions of the feature map are concurrently reduced from \(32 \times 32\) to \(16 \times 16\), \(8 \times 8\), \(4 \times 4\), and ultimately \(2 \times 2\).

An adaptive average pooling layer then compresses the spatial dimensions to a single pixel while preserving the 64-channel depth. Passing the resulting \(1 \times 1 \times 64\) tensor through a fully connected linear layer maps it to a scalar output. A sigmoid activation function is finally applied to produce the final prediction in the $[0,1]$ range.




\end{appendices}


\end{document}